\documentclass{IEEEoj}
\usepackage{cite}
\usepackage{amsmath,amssymb,amsfonts}
\usepackage{amsthm}
\usepackage{algorithmic}
\usepackage{graphicx,color}
\usepackage{textcomp}
\usepackage{subcaption}
\usepackage{multirow}
\newtheorem{definition}{Definition}
\usepackage{arydshln}
\usepackage{soul}
\usepackage{dsfont}
\usepackage{url}
\usepackage{tcolorbox}
\usepackage{dblfloatfix}
\usepackage{hyperref}

\def\BibTeX{{\rm B\kern-.05em{\sc i\kern-.025em b}\kern-.08em
    T\kern-.1667em\lower.7ex\hbox{E}\kern-.125emX}}
\AtBeginDocument{\definecolor{ojcolor}{cmyk}{0.93,0.59,0.15,0.02}}
\def\OJlogo{\vspace{-14pt}\includegraphics[height=28pt]{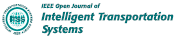}}
\begin{document}
\receiveddate{XX Month, XXXX}
\reviseddate{XX Month, XXXX}
\accepteddate{XX Month, XXXX}
\publisheddate{XX Month, XXXX}
\currentdate{XX Month, XXXX}
\doiinfo{OJITS.2022.1234567}

\title{Multi-hop Upstream Anticipatory Traffic Signal Control with Deep Reinforcement Learning}

\author{Xiaocan Li\authorrefmark{1}, Xiaoyu Wang\authorrefmark{2}, Ilia Smirnov\authorrefmark{2}, Scott Sanner\authorrefmark{1}, Baher Abdulhai\authorrefmark{2} }
\affil{Department of Mechanical \& Industrial Engineering, University of Toronto, Toronto, Canada}
\affil{Department of Civil \& Mineral Engineering, University of Toronto, Toronto, Canada}
\corresp{CORRESPONDING AUTHOR: Xiaocan Li (e-mail: hsiaotsan.li@mail.utoronto.ca).}
\markboth{Multi-hop Upstream Anticipatory Traffic Signal Control with Deep Reinforcement Learning}{Li \textit{et al.}}

\begin{abstract}
Coordination in traffic signal control is crucial for managing congestion in urban networks. Existing pressure-based control methods focus only on immediate upstream links, leading to suboptimal green time allocation and increased network delays. However, effective signal control inherently requires coordination across a broader spatial scope, as the effect of upstream traffic should influence signal control decisions at downstream intersections, impacting a large area in the traffic network. Although agent communication using neural network-based feature extraction can implicitly enhance spatial awareness, it significantly increases the learning complexity, adding an additional layer of difficulty to the challenging task of control in deep reinforcement learning. To address the issue of learning complexity and myopic traffic pressure definition, our work introduces a novel concept based on Markov chain theory, namely \textit{multi-hop upstream pressure}, which generalizes the conventional pressure to account for traffic conditions beyond the immediate upstream links. This farsighted and compact metric informs the deep reinforcement learning agent to preemptively clear the multi-hop upstream queues, guiding the agent to optimize signal timings with a broader spatial awareness. Simulations on synthetic and realistic (Toronto) scenarios demonstrate controllers utilizing multi-hop upstream pressure significantly reduce overall network delay by prioritizing traffic movements based on a broader understanding of upstream congestion.
\end{abstract}



\maketitle

\section{INTRODUCTION}
\IEEEPARstart{T}{raffic} signal control (TSC) is a cornerstone of intelligent transportation systems, designed to optimize traffic flow at intersections, reduce congestion, and minimize delays. Traditional methods, such as pre-timed and actuated control, have been widely adopted \cite{webster1958traffic, warberg2008greenwave, little1981maxband, hunt1982scoot}, but they often struggle to adapt to dynamic and complex traffic conditions. To address these limitations, the concept of traffic pressure has emerged as a promising metric for adaptive signal control strategies. Traffic pressure, at the intersection level, quantifies the disparity in traffic statistics (e.g., vehicle count or density) between upstream and downstream links \cite{varaiya2013maxpressure, varaiya2013maxpressure_springer}, enabling more responsive control approaches \cite{tsitsokas2023two}. For instance, PressLight \cite{wei2019presslight} integrated traffic pressure into reinforcement learning (RL) agent's reward design to improve network efficiency. 

\begin{figure}[htbp]
\centering
\includegraphics[width=\columnwidth]{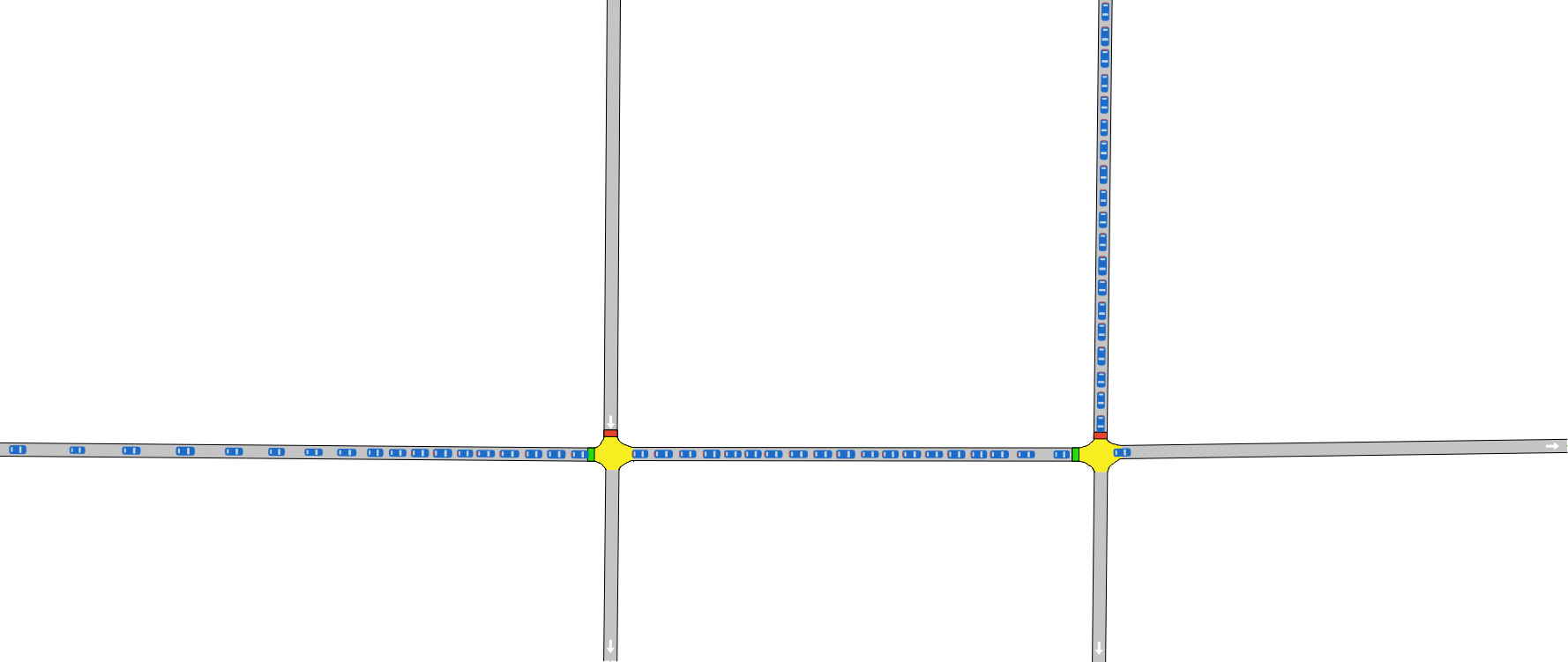}
\caption{A motivating example to demonstrate the issue of existing myopic pressure definition and the need for farsighted multi-hop upstream pressure definition.}
\label{fig:motivating-example}
\end{figure}

Despite these advancements, existing traffic pressure metrics remain limited in their spatial scope, focusing solely on immediate links at individual intersections while ignoring the broader network context. 
As the minimal motivating example shown in Figure \ref{fig:motivating-example}, at the right intersection, a myopic pressure-based controller would assign equal green time to eastbound and southbound flows because it perceives equal pressures from immediate upstream links. This approach neglects the longer queues and accumulating congestion further upstream on the eastbound route. This example is verified in numerical experiments in Section \ref{sec:exp-setup}. Such myopic decision-making exacerbates delays and reduces overall network efficiency, highlighting the need for a farsighted metric that accounts for multi-hop upstream conditions.

The goal of this work is to develop a generalized concept of traffic pressure that integrates \textit{multi-hop upstream} conditions. This approach captures a more comprehensive view of traffic dynamics, allowing controllers to prioritize traffic movements that most effectively alleviate congestion. By integrating multi-hop upstream pressure into deep RL agent design, this work provides a farsighted and adaptive framework that mitigates network delays and improves the overall performance of urban traffic networks.

The contribution of this work are three folds:
\begin{itemize}
\item \textit{Generalization of Traffic Pressure}: This paper introduces a novel concept of multi-hop upstream pressure grounded in Markov chain theory, which extends the conventional myopic traffic pressure to account for upstream conditions beyond immediate incoming links. This novel metric incorporates a broader spatial awareness than the traditional counterpart.
\item \textit{Integration into Deep Reinforcement Learning}: The multi-hop upstream pressure is integrated into the deep reinforcement learning framework, informing the agent's observation and reward spaces. This encourages preemptive queue clearance and more effective signal timing optimization based on upstream traffic conditions, addressing the limitations of existing RL controllers based on traditional pressure definition.
\item \textit{Comprehensive Validation}: The effectiveness of the proposed approach is validated through extensive simulations on both synthetic and realistic traffic scenarios, including a Toronto-based case study. Results show significant reductions in overall network delays, demonstrating the practical advantages of using multi-hop upstream pressure for traffic signal control.
\end{itemize}

\section{LITERATURE REVIEW}
In this section, we provide literature review on traffic signal control, and the variations of traffic pressures and their applications.

\subsection{Multi-intersection Traffic Signal Coordination}

\paragraph{Traditional Traffic Signal Control}

Traditional traffic signal control methods primarily focus on signal progression to optimize traffic flow. Pre-timed approaches, such as GreenWave \cite{warberg2008greenwave} and Maxband \cite{little1981maxband}, synchronize offsets across intersections to reduce vehicle stops in specific directions. While effective for stable traffic patterns, these methods lack adaptability to dynamic conditions.

Actuated and classical adaptive systems enhance real-time flexibility. Actuated control adjusts signals based on immediate traffic detection, but its myopic nature limits network-level coordination \cite{Ni2020actuated}. Adaptive systems like SCATS \cite{lowrie1990scats} and SCOOT \cite{hunt1982scoot} expand coordination regionally or hierarchically but rely on pre-designed models, reducing their effectiveness in highly dynamic environments.

\paragraph{Reinforcement Learning  based Adaptive Control}

Reinforcement Learning (RL) has emerged as a promising approach for adaptive traffic signal control, leveraging data-driven techniques to optimize signal timings dynamically. RL-based methods are categorized into centralized, hierarchical, and decentralized structures.
\begin{itemize}
    \item \textit{Centralized Control}: A single agent observes and controls the entire network \cite{kuyer2008multiagent}. While this approach achieves network-level coordination, it struggles with scalability in large networks.

    \item \textit{Hierarchical Control}: Multiple levels of agents are deployed, with upper-level agents providing macroscopic instructions and lower-level agents making finer decisions \cite{geng2024hisoma, zhou2024cooperative, xu2021hierarchically}. Hierarchical control balances scalability with coordination but requires careful design to ensure smooth interaction between agent levels.

    \item \textit{Decentralized Control}: Each intersection is controlled by an independent agent \cite{wei2019presslight}, making the system highly scalable. However, the lack of inherent coordination can lead to suboptimal network performance. Strategies to enhance coordination include:
    \begin{itemize}
        \item \textit{Centralized Training with Decentralized Execution (CTDE)}: This approach trains agents jointly using shared information while maintaining decentralized execution during deployment \cite{el2013multiagent}.
        \item \textit{Agent Communication}: Communication frameworks allow agents to exchange local traffic states and coordinate actions, improving global performance. Neighbor RL \cite{arel2010reinforcement} directly concatenates immediate neighbor intersections' information. GCNRL \cite{nishi2018traffic} uses Graph Convolutional Networks (GCN) to extract features across intersections. CoLight \cite{wei2019colight} leverages Graph Attentional Networks (GAT) to facilitate communication. eMARLIN \cite{wang2024emarlin, wang2023emarlinplus} embeds immediate neighbor intersections information into an embedding space. The reward designs of all these methods are only associated with local intersections, making these agents less farsighted.
    \end{itemize}
\end{itemize}

Effective decentralized RL for TSC relies heavily on the design of agent observations and rewards, as the information available to each agent directly impacts its ability to make informed decisions. For example, PressLight \cite{wei2019presslight} integrated traffic pressure into reward design. However, the vanilla traffic pressure is limited in its scope to immediate neighbors, while conditions beyond immediate links are critical for intersection coordination. In addition, feature extraction via neural networks for agent communication \cite{wei2019colight, wang2024emarlin} imposes additional computational overhead and learning difficulties that further complicate the control task based on deep reinforcement learning. \textit{This motivates the exploration of efficient and effective observation and reward designs that capture broader traffic conditions.} Our approach complements existing agent communication frameworks that extract shared information via neural networks.

\subsection{Traffic Pressure and Its Variations}


The traffic pressure concept originates from resource reallocation strategies in wireless communication networks \cite{tassiulas1990stability}. The primary application of traffic pressure is the MaxPressure control policy, which determines phase activation \cite{wongpiromsarn2012distributed, varaiya2013maxpressure, varaiya2013maxpressure_springer, zaidi2015traffic, wu2017delay} or green time allocation \cite{kouvelas2014maximum, le2015decentralized, tsitsokas2023two, mercader2020max} in decentralized traffic control systems. MaxPressure has also been integrated into perimeter control strategies \cite{tsitsokas2023two, liu2024nmp} that prevent regional congestion by restricting the inflow to protected regions. While effective, some implementations of MaxPressure with phase activation-based action spaces have raised concerns about confusing phase sequences, which could frustrate drivers and increase safety risks. Solutions include fixed or variable cycle times and predefined phase orders, combined with stability guarantees \cite{le2015decentralized, levin2020max}. Further enhancements include integrating vehicle rerouting into MaxPressure for improved performance \cite{zaidi2015traffic}.

\paragraph{Variations in Traffic Pressure Definition}
Numerous variations of traffic pressure have been developed, focusing on specific traffic statistics:
\begin{itemize}
    \item \textit{Queue Density}: Incorporating link lengths into pressure calculation ensures that shorter links with queues are prioritized over longer links with the same queue length \cite{kouvelas2014maximum}. This pressure definition is also used in reward design for RL-based signal control \cite{wei2019presslight}.
    \item \textit{Phase Weights}: To prioritize specific phases, dynamic weights are introduced \cite{xiao2015further, xiao2015throughput}, along with adaptive estimation of turning ratios and saturation flows.
    \item \textit{Delay Time}: To improve fairness in waiting times, traffic delay is included in pressure definitions \cite{wu2017delay, liu2022dmp}.
    \item \textit{Travel Time}: Recognizing the difficulty in measuring queues, travel times have been used as proxies for pressure definitions and tested in both simulation and real-world settings \cite{mercader2020max}.
    \item \textit{Platoon and Occupancy Prioritization}: C-MP incorporates space mean speed to prioritize large moving platoons \cite{ahmed2024cmp}, OCC-MP prioritizes high-occupancy vehicles to improve passenger-based efficiency \cite{ahmed2024occ}, and PQ-MP integrates pedestrian queues to account for mixed traffic scenarios \cite{xu2024pedmp}.
\end{itemize}

\paragraph{Multi-hop Extensions}
Traffic pressure has been extended to multi-hop downstream applications for perimeter control. For instance, N-MP deprioritizes phases when multi-hop downstream link densities exceed a critical threshold \cite{liu2024nmp}, and \cite{li2024generalized} formalizes multi-hop downstream pressure grounded on Markov chain theory. However, these approaches primarily focus on downstream conditions, neglecting upstream traffic dynamics.


Capturing the potential of \textit{upstream} traffic is crucial for preemptively clearing queues. To the best of our knowledge, existing pressure-based works only consider immediate upstream links, without extending the pressure's scope to further upstream conditions. To address the limitation of upstream scope, this work introduces a novel concept of \textit{multi-hop upstream pressure}, grounded in Markov chain theory. The proposed metric is integrated into the observation space and reward function of deep reinforcement learning agents, enabling preemptive signal timing optimization and effective coordination across intersections.

\section{PROBLEM STATEMENT}

\subsection{Traffic Signal Control as Decentralized Markov Decision Processes}
In this work, traffic signal control is modeled as a Decentralized Markov Decision Process (DecMDP), which is defined by the tuple $(n, \mathcal{S}, \mathcal{A}, \mathcal{T}, \mathcal{R}, \gamma)$, where $\mathcal{S}=\cup_{i=1}^{n} \mathcal{O}_i$ is the system state space as joint observation spaces of $n$ agents, and $\mathcal{A}=\cup_{i=1}^{n} \mathcal{A}_i$ represents the joint action of $n$ agents:

\begin{itemize}
    \item \textbf{Observation space} $\mathcal{O}_i$: For each intersection $i$ controlled by agent $i$, the observation space consists of the multi-hop pressure associated with each phase. For instance, if an intersection has an eastbound phase and a southbound phase, then the observation space is two-dimensional, representing the pressure for the eastbound and southbound phases, respectively. Mathematical definition for observation space can be found in Section \ref{sec:method}-\ref{sec:obs-design}.
    \item \textbf{Action space} $\mathcal{A}_i$: For each intersection $i$ controlled by agent $i$, the action is defined as the cycle splits, representing the proportion of green time allocated to each phase at the intersection.
    \item \textbf{Transition probability} $\mathcal{T}$: The transition probability $\mathcal{T}(s'|s, a)$ represents the probability of transitioning from the current global state $s$ to a new global state $s'$ after the joint action $a = (a_1, a_2, ..., a_n)$ is taken by the $n$ agents. This probability models the dynamics of traffic flow in response to changes in signal timings. In our setting, $\mathcal{T}$ is handled by the traffic simulator and is not exposed to agents.
    \item \textbf{Reward} $\mathcal{R}_i$: For each intersection $i$ controlled by agent $i$, the reward is calculated as the negative sum of multi-hop potentials over phases. Mathematical definition for reward space can be found in Section \ref{sec:method}-\ref{sec:reward-design}. This reward design encourages each agent to clear upstream traffic as quickly as possible. It is important to note that a myopic pressure-based reward may lead to undesirable behavior, such as holding vehicles upstream to minimize the myopic pressure.
    \item \textbf{Discount factor} $\gamma$: The discount factor $\gamma \in [0, 1]$ determines the relative importance of future rewards compared to immediate rewards.
\end{itemize}

Reinforcement learning is employed to solve this MDP by training each agent $i$ to learn an optimal policy $\pi_i$, which maximizes the expected discounted cumulative reward $\mathbb{E}_{a_i \sim \pi_i}\left[ \sum_{k=0}^{\infty} \gamma^k R_i (o_{i}^{(t+k)}, a_{i}^{(t+k)})\right]$. Through repeated interactions with the environment, each agent observes the traffic conditions, selects actions, receives rewards, and updates its policy to improve long-term traffic efficiency.

In this study, we utilize the Proximal Policy Optimization (PPO) algorithm \cite{schulman2017proximal}, a widely used RL algorithm known for its stability and efficiency.

\section{METHODOLOGY}\label{sec:method}
This section outlines the framework for implementing a generalized multi-hop pressure model in traffic signal control. We first clearly define the mathematical notations that are used throughout this work in Table \ref{tab:math-definitions}. Then, we model the traffic network structure with graph representations. Finally, we define and illustrate the multi-hop pressure and multi-hop potential metrics, both in its scalar and vectorized forms, and demonstrate its calculation through a simplified example.

\begin{table*}[htbp]
\centering
\caption{Mathematical notations.}
\begin{tabular}{p{0.3\columnwidth}p{1.6\columnwidth}}
\hline
\textbf{Symbol} & \textbf{Definition} \\ \hline
\multicolumn{2}{l}{\textbf{Graph Representation Related Notations}} \\
$\mathcal{V}$ & The set of the whole traffic network's links. This is also the set of graph vertices. \\ 
$\mathcal{E}$ & The set of graph edges. Each edge represents a permissible turning movement. \\ 
$\Omega$ & Supersink, an abstract node that merges all destinations. \\ 
$\mathcal{V}^e$ & The extended set of traffic links, including the supersink compared to $\mathcal{V}$. See Definition \ref{def:extended-graph}. \\ 
$\mathcal{E}^e$ & The extended set of edges. See Definition \ref{def:extended-graph}. \\ 
$G = (\mathcal{V}, \mathcal{E})$ & The graph representation of the traffic network without supersink. \\ 
$G^e = (\mathcal{V}^e, \mathcal{E}^e)$ & The graph representation of the traffic network with supersink. See Definition \ref{def:extended-graph}. \\ 
$\mathcal{N}_u (l,h)$ & The set of $h$-hop upstream links from link $l$. 0-hop means the link itself, i.e., $\mathcal{N}_u (l,0)=\{l\}$. \\ 
$\mathcal{N}_d (l,h)$ & The set of $h$-hop downstream links from link $l$. 0-hop means the link itself, i.e., $\mathcal{N}_d (l,0)=\{l\}$. \\ \hline
\multicolumn{2}{l}{\textbf{Pressure \& Potential Related Notations}} \\ 
$T_{ij}$ [unitless] & Turning ratio from link $i$ to $j$. The sum of turning ratios from link $i$ to all its 1-hop downstream links must be 1: \newline 
$\sum_{j\in\mathcal{N}_d(i,1)} T_{ij} = 1 \quad \forall i$ and $0 \leq T_{ij} \leq 1 \quad \forall i,j$  \\ 
$\mathbf{T}\in \mathbb{R}^{|\mathcal{V}|\times|\mathcal{V}|}$ & The weighted adjacency matrix of graph $G$ where the ($i,j$)-entry is $T_{ij}$. \\ 
$\mathbf{P}\in \mathbb{R}^{|\mathcal{V}^e|\times|\mathcal{V}^e|}$ & The weighted adjacency matrix of graph $G^e$, which is also a Markov transition matrix. See Eq. (\ref{eq:markov-chain-matrix}) for details. \\ 
$Q(l)$ [$veh$] & Queue length of link $l$. Default speed threshold in simulator: Queue entering: 2$m/s$, Queue exit: 4$m/s$. \\ 
$\mathbf{Q}\in\mathbb{R}^{|\mathcal{V}^e|}$ & The concatenation of queue lengths for links in $\mathcal{V}^e$. The order matches the rows and columns of $\mathbf{P}$. \\ 
$p(l,h)$ [$veh/km$] & The pressure with $h$-hop upstream for link $l$. \\ 
$\mathbf{p}(h)\in\mathbb{R}^{|\mathcal{V}^e|}$ & The concatenation of $h$-hop pressure for all links. The arranging order matches that of $\mathbf{Q}$. \\ 
$\phi(l,h)$ [$veh/km$] & The potential with $h$-hop upstream for link $l$. \\ 
$\mathbf{\Phi}^{\text{up}}(h)$ [$veh/km$] & The concatenation of $h$-hop potential for all links. The arranging order matches that of $\mathbf{Q}$. \\ 
$\mathbf{\Phi}^{\text{down}}$ [$veh/km$] & The concatenation of immediate downstream traffic potential for all links. The arranging order matches that of $\mathbf{Q}$. \\ 
$L_{\text{in}}(i)$ & The set of incoming links for intersection $i$. \\ 
$L_{\text{in}}(i, \theta)$ & The set of incoming links for phase $\theta$ in intersection $i$. \\ 
$\Theta (i)$ & The set phases in intersection $i$ controlled by RL. \\ 
$p(\theta)$ [$veh/km$] & Phase pressure for phase $\theta$. See Definition \ref{def:phase-pressure}. \\ \hline
\end{tabular}
\label{tab:math-definitions}
\end{table*}

\subsection{Graph Representations of Traffic Networks}\label{sec:graph}

The traffic network is represented as a graph described in Definition \ref{def:extended-graph}. To simplify this representation, a supersink $\Omega$ is introduced, consolidating all destinations into a single abstract node. Incorporating the supersink allows the adjacency matrix to exhibit the properties of a Markov chain transition matrix, enabling mathematical operations on the adjacency matrix to be interpreted through the Markov chain theory. The supersink is characterized by the following properties:
\begin{itemize} 
\item \textit{Zero Queue Density}: With infinite capacity, the supersink's queue density is always zero. 
\item \textit{Absorption}: Links connected to the supersink are fully absorbed, and the supersink remains its own downstream neighbor. This property establishes vehicle movement on the graph as an Absorbing Markov Chain. 
\item \textit{Binary Turning Ratio}: The turning ratio for any link connected to the supersink or for transitions within the supersink itself is 1, and 0 for all other cases. 
\end{itemize}

\begin{definition} [Graph representation]\label{def:extended-graph}
    The graph representation $G^e = (\mathcal{V}^e, \mathcal{E}^e)$, where:
    \begin{itemize}
    \item The extended link set $\mathcal{V}^e$ additionally includes a supersink vertex $\Omega$, i.e., $\mathcal{V}^e = \mathcal{V}\cup\{\Omega\}$.
    \item The extended edge set $\mathcal{E}^e$ additionally includes those edges reflecting connections to the supersink.
    \item Edge weight $T_{uv}$ is the turning ratio from link $u$ to link $v$. These weights are derived from real empirical data or traffic simulations, representing the probability of traffic flow transitions between links.
    \end{itemize}
\end{definition}

\begin{figure}[htbp]
\centering
  \begin{subfigure}[b]{0.48\columnwidth}
    \includegraphics[width=\linewidth]{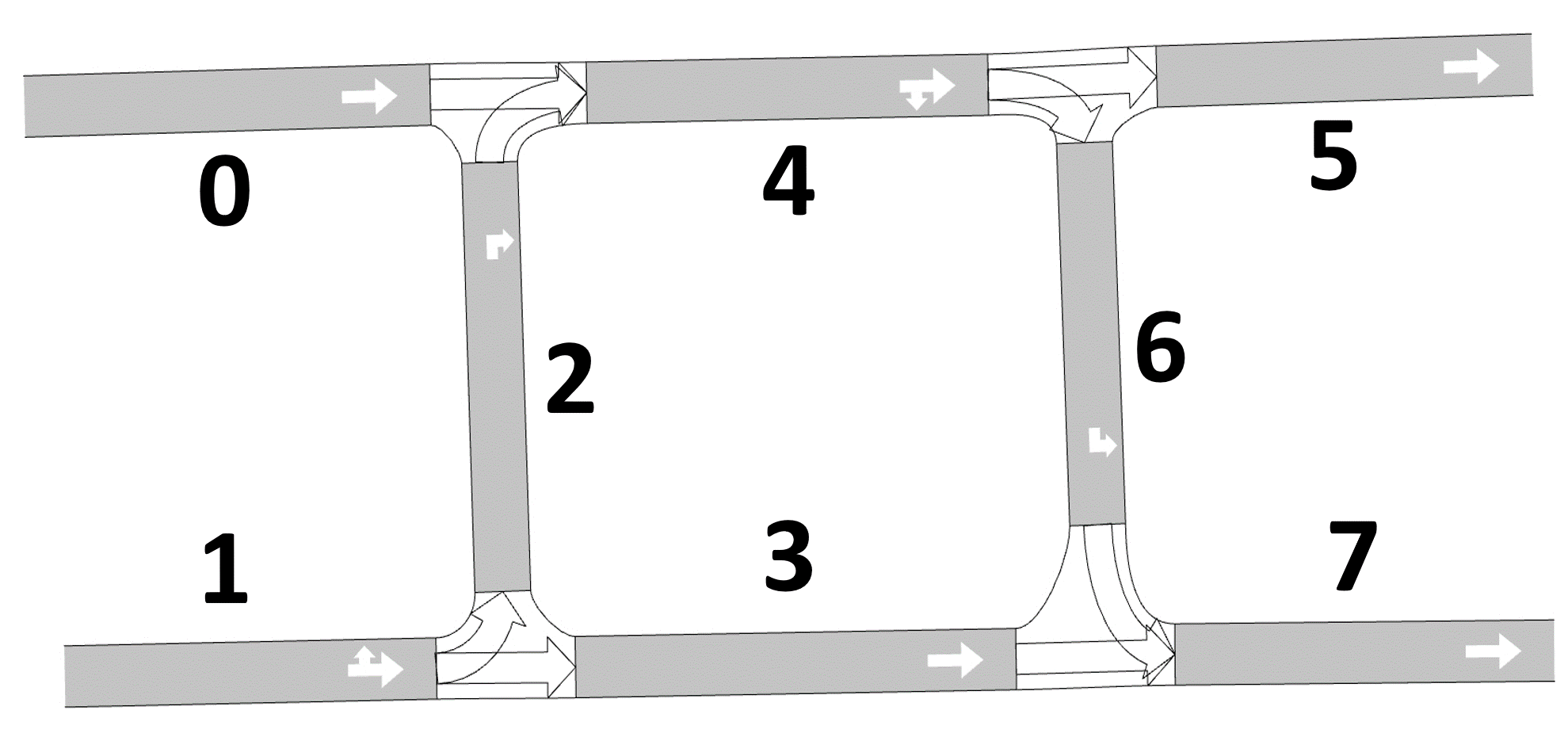}
    \caption{A toy network with 8 links.}
    \label{fig:toy-network}
  \end{subfigure}
  ~
  \begin{subfigure}[b]{0.48\columnwidth}
    \includegraphics[width=\linewidth]{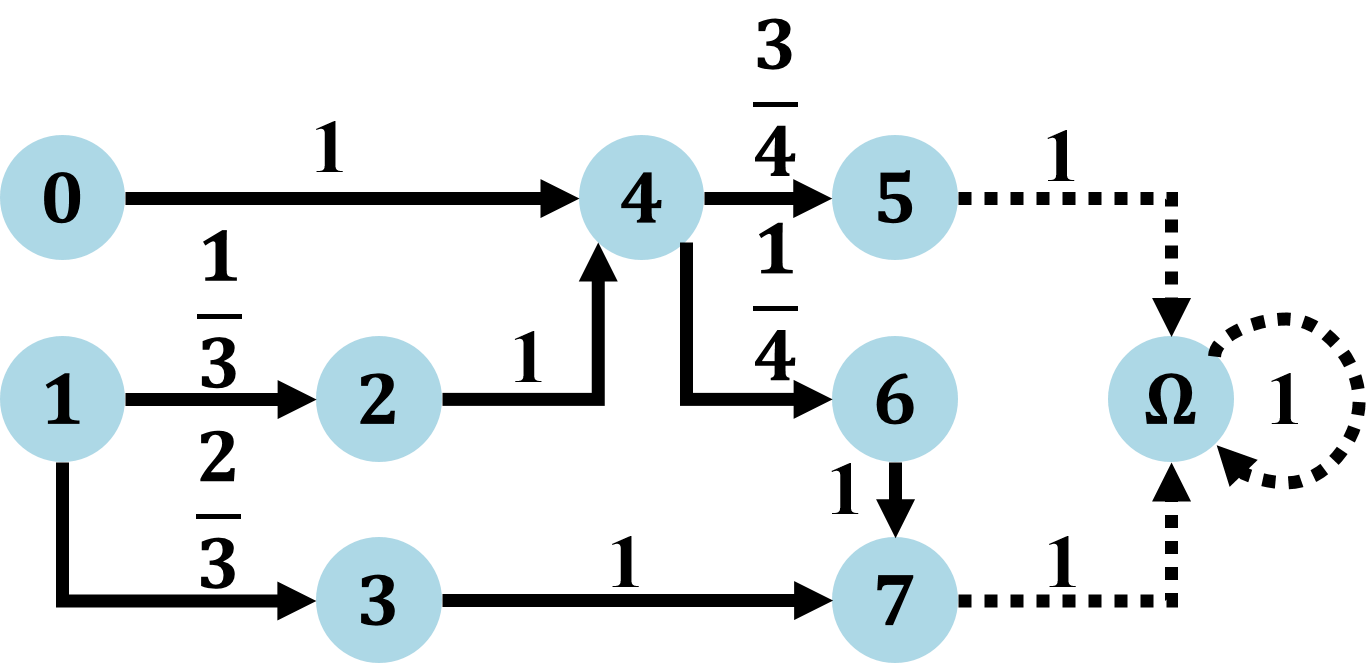}
    \caption{Graph representation of the toy network.}
    \label{fig:graph-representation-toy-network}
  \end{subfigure}
\caption{An example of graph representation for a toy traffic network. (a) A traffic network with 8 traffic links indexed from 0 to 7. (b) The weights shown on the edges are the fabricated turning ratios. The vertex $\Omega$ is the supersink, and the edges in  dashed lines represent graph $G^e$ being extended from graph $G$.}
\end{figure}

To assist understanding of graph representations, an example is provided with a simplistic traffic network with 8 links in Figure \ref{fig:toy-network}, and is mapped onto its graph representation depicted in Figure \ref{fig:graph-representation-toy-network}, where the turning ratios are labeled on edges.

\subsection{Vehicle Movement as an Absorbing Markov Chain}

The movement of a vehicle within a traffic network, guided by specific turning ratios, can be modeled as a time-homogeneous absorbing Markov chain. In this model, the presence of a vehicle on a link $l$ is represented as a random variable with probability $Pr(x=l)$. The state space of the Markov chain is finite, comprising $|\mathcal{V}^e|$ states, corresponding to the total number of links in the network. The transition matrix $\mathbf{P}$ corresponds to the weighted adjacency matrix of the graph $\mathcal{G}^e$, defined as follows:
\begin{align}\label{eq:markov-chain-matrix}
    \mathbf{P} = \left[\begin{array}{ccc:c}
        T_{11} & \ldots & T_{1|\mathcal{V}|} & T_{1\Omega}\\
        \vdots & \ddots & \vdots & \vdots \\
        T_{|\mathcal{V}|1} & \ldots & T_{|\mathcal{V}||\mathcal{V}|} & T_{|\mathcal{V}|\Omega}\\
        \hdashline
        0      & \ldots & 0      & 1
    \end{array}\right] = 
    \begin{bmatrix}
        \mathbf{T}               & \mathbf{T}_{*\Omega} \\
        \mathbf{0}^\top & 1
    \end{bmatrix},
\end{align}


\subsection{Multi-hop Upstream Pressure: A Customizable Metric for Far-reaching Upstream Traffic Condition}
Congestion from immediate upstream links has a more direct and significant impact than congestion several blocks further upstream. Therefore, the multi-hop pressure definition needs to capture the \textit{diminishing influence} of distant congestion while still accounting for its \textit{cumulative effect} on the current traffic link. To understand the upstream links at higher hops, we provide an example of upstream links for link 7. Mathematically, they are written as:
\begin{align}
    \mathcal{N}_u (7, 0) &= \{7\}\\
    \mathcal{N}_u (7, 1) &= \{3, 6\}\\
    \mathcal{N}_u (7, 2) &= \{1, 4\}\\
    \mathcal{N}_u (7, 3) &= \{0\} \\
    \mathcal{N}_u (7, h) &= \{\}, \quad h\geq 4, h\in\mathbb{N}^+
\end{align}

The scalar formulation of multi-hop upstream pressure, designed to calculate the pressure for a single link, is less compact and is thus presented in the Appendix. In contrast, the vectorized formulation enables simultaneous computation of pressures for \textit{all} links in the traffic network. This vectorized approach significantly improves computational efficiency compared to processing each link individually.

\subsubsection{Vectorized Version: Multi-hop Pressure for All Links}


Unlike the scalar version could only compute pressure for one link at a time, the vectorized version can compute pressures for \textit{all} links in the traffic network simultaneously, which accelerates the computation upon implementation compared to iterating over each link in the traffic network:
\begin{tcolorbox}[colback=lightgray!20, colframe=black, sharp corners]
\begin{align}
    &\text{Recursive Form:} \nonumber \\
    &\mathbf{p}(0) = \mathbf{Q} - \mathbf{PQ}  \label{eq:markov-0-hop} \\
    &\mathbf{p}(h) = \mathbf{p}(h-1) + (\mathbf{P}^{h})^\top\mathbf{Q}, \quad h\in\mathbb{N}^+   \label{eq:markov-recursive-matrix-pressure} \\
    & \nonumber \\
    &\text{Unrolled Form:} \nonumber \\
    &\mathbf{p}(h) =\sum_{h'=0}^{h} (\mathbf{P}^{h'})^\top \mathbf{Q} - \mathbf{PQ}, \quad h\in\mathbb{N} \label{eq:markov-unroll-matrix-pressure}
\end{align}
\end{tcolorbox}

\paragraph{Interpretations of $\mathbf{P}^h$} 

The term $\mathbf{P}^h$ in Eq. (\ref{eq:markov-recursive-matrix-pressure}) deserves meticulous interpretation. In Markov chain theory, the entry $(i,j)$ in the $h$-th power of the transition matrix $\mathbf{P}^h$, denoted as $(\mathbf{P}^h)_{ij}$, indicates the probability of transitioning from vertex $i$ to vertex $j$ in exactly $h$ steps. In the context of a traffic network, where entries correspond to turning ratios, $(\mathbf{P}^h)_{ij}$ represents the probability of a vehicle traveling from link $i$ to link $j$ through any possible sequence of $h$ links. This implies that link $i$ is one of the $h$-hop upstream links of link $j$, i.e., $i \in \mathcal{N}_u (j, h)$.

\begin{itemize} 

\item $h$\textit{-hop influence}: The entry $(\mathbf{P}^h)_{ij}$ quantifies the influence of link $i$ on link $j$ after $h$ transitions. It reflects the notion that traffic pressure propagates across the network, extending beyond local effects to distant links. 

\item \textit{Decay of influence over hop}: As $\mathbf{P}^h$ involves repeated multiplication of $\mathbf{P}$, the influence decreases with increasing $h$ due to turning ratios being bounded within $[0, 1]$. This captures the natural attenuation of congestion effects over distance in a traffic network. 

\item \textit{Pressure contribution}: Multiplying $[(\mathbf{P}^h)^\top]_{:, j}$ by $\mathbf{Q}$ takes weighted sums on the traffic condition (e.g., queue density) over all $h$-hop upstream links exerted on link $j$. High congestion at link $i$, combined with a significant $[(\mathbf{P}^h)^\top]_{ij}$, results in a substantial contribution to the pressure at link $j$. The cumulative contribution over all 0 to $h$ hops upstream links, formalized as \textit{h-hop upstream potential}, is discussed in Section \ref{sec:method}-\ref{sec:reward-design}. 

\item \textit{Independent contribution}: The term $(\mathbf{P}^h)^\top \mathbf{Q}$ captures the additional pressure exerted on a link by queues at $h$-hop upstream links, not included in $(h-1)$-hop upstream links. This isolates the unique contribution of the $h$-hop and highlights how congestion propagates spatially and temporally. Understanding this distinction is critical for designing controllers that mitigate congestion effectively using multi-hop pressure information. 
\end{itemize}


\subsection{Observation Space Design: Multi-hop Upstream Pressure for Phases}\label{sec:obs-design}
\begin{definition} [Phase Pressure] \label{def:phase-pressure}
    Given a control plan for an intersection $i$, the phase pressure is defined as the summation of link pressure over all incoming links in  phase $\theta$: 
    \begin{align}
        p(\theta) &= \sum_{l \in L_{\text{in}}(i, \theta)} p(l, h)
    \end{align}
\end{definition}

The observation $o_i \in \mathcal{O}_i$ for agent $i$ is the concatenation of phase pressure in intersection $i$:
\begin{align}
    o_i = \|_{\theta \in \Theta (i)} \quad p(\theta)
\end{align}

\subsection{Reward Design: Multi-hop Upstream Potentials}\label{sec:reward-design}
Adopted from physics, another perspective for the pressure in Eq. (\ref{eq:markov-unroll-matrix-pressure}) is the difference of upstream potential and downstream potential:
\begin{align}
    \mathbf{p}(h) &= \mathbf{\Phi}^{\text{up}} (h) - \mathbf{\Phi}^{\text{down}}
\end{align}
where $\mathbf{\Phi}^{\text{up}} (h)$ is the $h$-hop upstream traffic potential and $\mathbf{\Phi}^{\text{down}}$ is the immediate downstream traffic potential:
\begin{align}
    \mathbf{\Phi}^{\text{up}} (h) &= \sum_{h'=0}^{h}(\mathbf{P}^{h'})^\top \mathbf{Q} \\
    \mathbf{\Phi}^{\text{down}} &= \mathbf{PQ}
\end{align}

To encourage RL agents to clear upstream queues, the reward for agent $i$ is defined as the negation of the $h$-hop upstream potential across all incoming links at intersection $i$:
\begin{align}
    r_i = - \sum_{l\in L_{\text{in}}(i)} \phi^{\text{up}}(l, h) \label{eq:potential-reward}
\end{align}

A higher number of upstream hops in the reward calculation encourages the agent to preemptively allocate longer green times to clear upstream queues, facilitating coordination across intersections. Notably, the number of hops used for observation is the same as that used in the reward.

\textit{Difference to pressure-based reward}: For comparison purpose, we also provide pressure-based reward formalization as the negation of the $h$-hop upstream potential across all incoming links at intersection $i$:
\begin{align}
    r_i = - \sum_{l\in L_{\text{in}}(i)} p(l, h) \label{eq:pressure-reward}
\end{align}

When $h=0$, it is a special case of myopic pressure reward used in PressLight \cite{wei2019presslight}.

\section{EXPERIMENTAL SETUP}\label{sec:exp-setup}

The proposed traffic signal control scheme is evaluated using the traffic simulator Aimsun \cite{Aimsun}. This section provides detailed description of the experimental setup in traffic network architecture and the traffic demand.

\subsection{Tested Scenarios}\label{sec:test-scenario}
Both synthetic and realistic scenarios are tested. When designing scenarios, we provide a wide range of complexity from the simplest scenario to a complicated one, to verify that our approach works on diverse scenarios.


\subsubsection{Synthetic Scenarios}
The scenario design breaks down into two parts: traffic network and traffic demand. We designed two traffic networks and three traffic demand saturation levels, resulting in $2\times  3 = 6$ synthetic scenarios in total.

\begin{figure}[htbp]
\centering
  \begin{subfigure}[b]{0.48\columnwidth}
    \includegraphics[width=\linewidth]{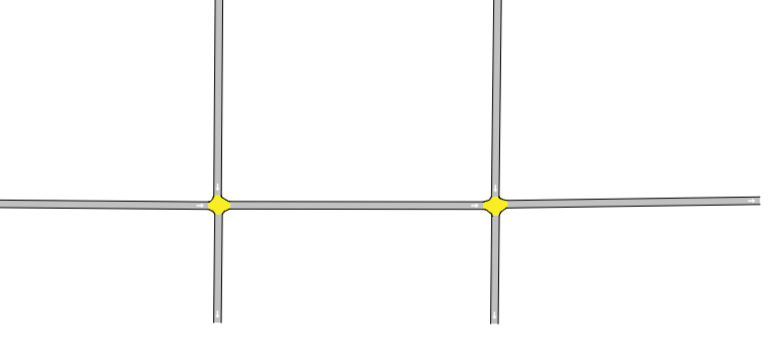}
    \caption{2-intersection  network.}
    \label{fig:network1x2}
  \end{subfigure}
  \hfill
  \begin{subfigure}[b]{0.48\columnwidth}
    \includegraphics[width=\linewidth]{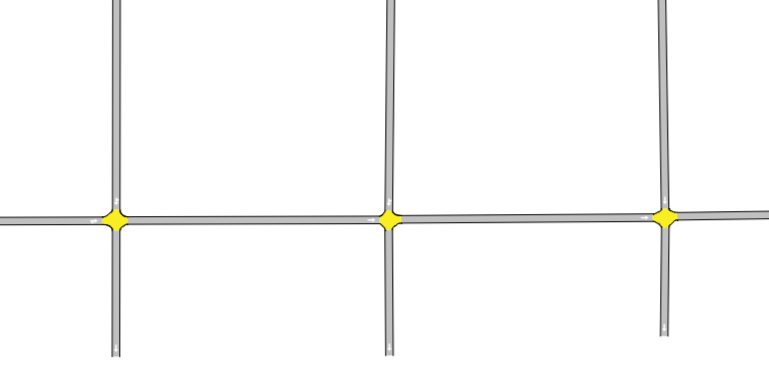}
    \caption{3-intersection  network.}
    \label{fig:network1x3}
  \end{subfigure}
\caption{Tested synthetic arterial networks. Link channelization and phasing scheme are described in Section \ref{sec:exp-setup}-\ref{sec:test-scenario}. }
\label{fig:synthetic-networks}
\end{figure}

\paragraph{Traffic Networks}
Two simplified traffic networks are synthesized, as shown in Figure \ref{fig:synthetic-networks}:
\begin{itemize}
    \item \textit{Network 1x2}: A minimal network with two intersections, designed to validate the effectiveness of multi-hop upstream pressure, adhering to the philosophy of minimal viable product in scientific research.
    \item \textit{Network 1x3}: An extension of Network 1x2, featuring three intersections along an arterial road.
\end{itemize}
    
\textit{Link channelization and phasing scheme}: Both synthetic networks share the following settings: The distance between adjacent intersections is 100 meters. Each link is single-lane and restricted to through movements, with no turning lanes.  All intersections are signalized, operating with two phases: eastbound movement and southbound movement. The cycle length is 90 seconds, including two 5-second interphases.

\paragraph{Traffic demands} The demand saturation level can be categorized into three levels in ascending order: 
\begin{itemize}
    \item \textit{Undersaturated}: 50\% of heavily saturated demand.
    \item \textit{Slightly saturated}: 75\% of heavily saturated demand.
    \item \textit{Heavily saturated}: The demand profile is tabulated in Table \ref{tab:demand-profile}, where the maximal traffic flow is 2700vph at the first 30 minutes, greatly exceeding the capacity of the intersection (approximately 1800vph). The southbound flow is at the rightmost intersection only.
\end{itemize}

\begin{table}[htbp]
\centering
\caption{Heavily saturated demand profile for synthetic networks. Flow unit: vph. Time unit: minute.}
\label{tab:demand-profile}
\begin{tabular}{cccccc}
\hline
Network                      & Direction & 0 - 30  & 30 - 60  & 60 - 90  & 90 - 120          \\ \hline
\multirow{2}{*}{Network 1x2} & EB        & 1800       & 0           & 0        & 0                    \\
                             & SB        & 900        & 0         & 0         & 0                    \\ \hline
\multirow{2}{*}{Network 1x3} & EB        & 1800       & 0           & 1000        & 0                    \\
                             & SB        & 900       & 900        & 900           & 0                    \\ \hline
\end{tabular}
\end{table}

One might question why a constant flow demand is not used in Network 1x3. The reason is that constant demand can be effectively managed by pre-timed constant control, thus a constant demand is insufficient to demonstrate the advantages of multi-hop upstream pressure. Instead, we design a dynamic demand that no constant controllers is optimal and highlighting the need for a more adaptive control.

For any of these 6 synthetic scenarios, the optimal controller is expected to allocate longer green times to the eastbound phase to accommodate greater EB demands at the rightmost intersection, while the other intersection(s) should consistently assign the maximum allowed green time to the eastbound phase, given the absence of southbound flow.

\subsubsection{Realistic Scenario}
\begin{figure*}[htbp]
\centering
\includegraphics[width=1.9\columnwidth]{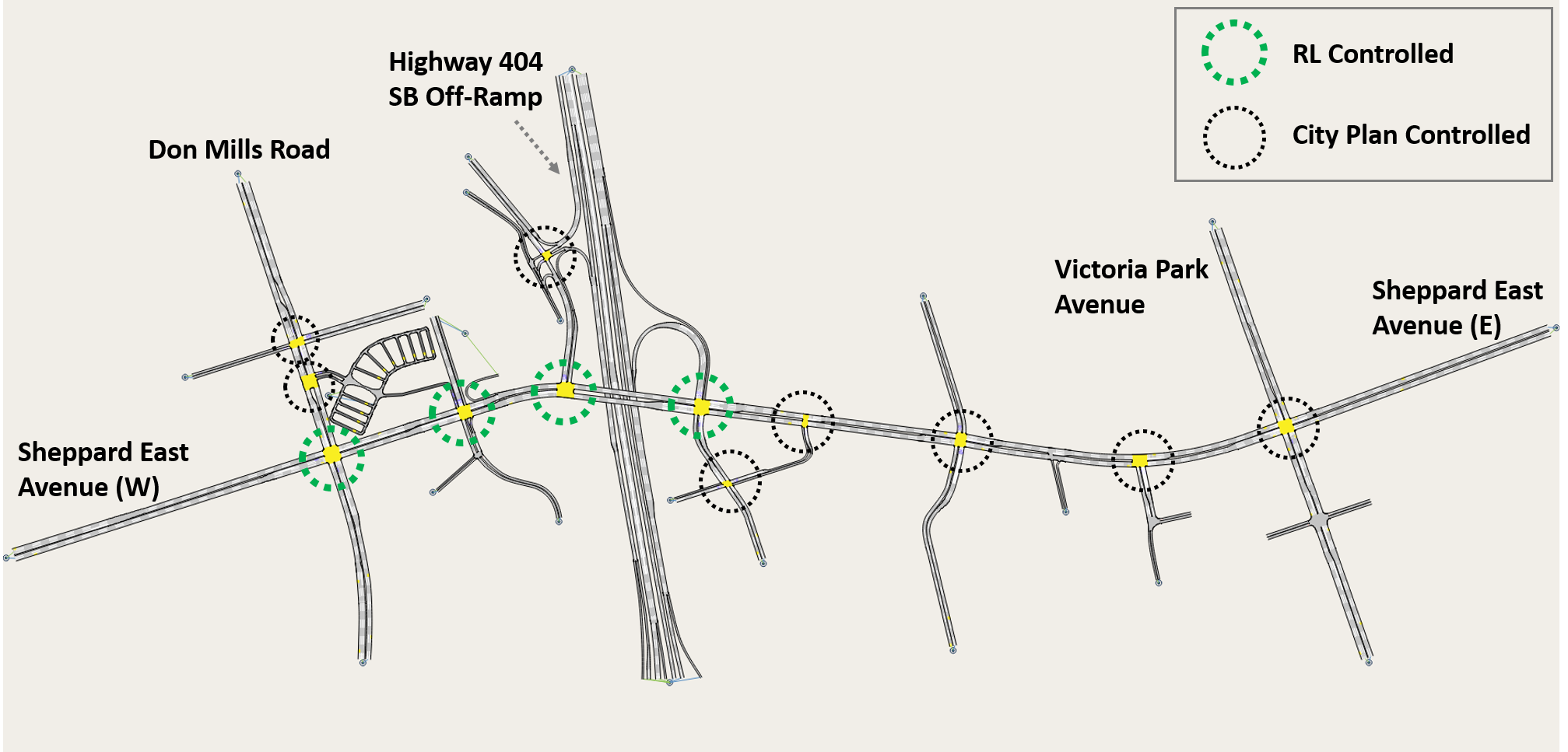}
\caption{The Toronto network testbed. Four consecutive intersections on the Sheppard Avenue corridor are controlled by our methods as they encounter large flows. The other eight intersections not experiencing heavy congestion are less critical, therefore are controlled by the city plan.}
\label{fig:sheppard-hwy404}
\end{figure*}

 \textbf{Toronto testbed}: This testbed simulates a neighborhood around the intersection of Sheppard Avenue and Highway 404 in Toronto, Ontario, Canada, implemented in the Aimsun simulator (Figure \ref{fig:sheppard-hwy404}). This neighborhood comprises $12$ signalized intersections, including the pivotal Sheppard Avenue intersection near a major bus station and the on/off ramps of Highway 404. Out of these 12 signalized intersections, 4 congestion-prone intersections on the Sheppard Avenue are controlled by our method, while the rest 8 signalized intersections experiencing light demand are controlled by the City Plan, as labeled in Figure \ref{fig:sheppard-hwy404}. The area also includes the Fairview Mall, which features large parking lots adjacent to the arterial road. Distances between consecutive signalized intersections range from $150$ to $300$ meters. The demand profile spans the morning period from 7:30 to 10:00 AM, reaching its peak around 9 AM. The demand consists of three types of vehicles: cars, trucks, and buses. The buses are simulated following the schedules of two public transit service providers: Toronto Transit Commission and York Region Transit. The demand is calibrated using publicly available turning movement counts from the City of Toronto. The linear regression coefficient  ($R^2=0.9119$) indicates a good fit of our demand profile to real-world traffic.



\paragraph{Baselines}
We compare our approach against established traffic control baselines. These baselines represent different commonly used traffic signal control strategies, ranging from non-adaptive pre-timed control to advanced learning-based adaptive methods. The following outlines the key baseline methods used in our evaluation:
\begin{itemize}
    \item \textit{Pre-timed (non-adaptive) Control: Webster method (Synthetic Scenarios Only)}: The cycle length and cycle splits are pre-defined according to historical flows.
    \item \textit{Learning-based Adaptive Control: PressLight (Both Synthetic Scenarios and Toronto Testbed)}: As a deep RL approach, PressLight leverages \textit{myopic} pressure in reward design, where the \textit{immediate} upstream and downstream traffic statistics are incorporated in pressure calculation.
    \item \textit{Semi-Actuated Control: City Plan (Toronto Testbed Only)}: A standard dual-ring NEMA phasing scheme with semi-actuated control \cite{holm2007traffic}. This control plan is replicated based on the actual implementation. One may request the traffic signal timing information from the City of Toronto \footnote{Request Signal Timing Information: \url{https://www.toronto.ca/services-payments/streets-parking-transportation/traffic-management/traffic-signals-street-signs/request-signal-timing-information/}}.
\end{itemize}

\paragraph{Evaluation Metrics} To comprehensively evaluate the performance of our proposed approach, we use three evaluation metrics that reflect different aspects of traffic statistics:
\begin{itemize}
    \item \textit{Total Time Spent (TTS) (hour)}: Summation of each vehicle's travel time spent starting from vehicle generation to exit, including time in virtual queue.
    \item \textit{Total Queue Time (Include Virtual) (hour)}: Summation of each vehicle's queue time from vehicle generation to exit. Therefore, time in the virtual queue is included.
    \item \textit{Total Virtual Queue Time (hour)}: Summation of each vehicle's time spent in the virtual queue.
\end{itemize}

\section{RESULTS \& DISCUSSION}

\subsection{Synthetic Scenario: Network 1x2}

\begin{figure}[htbp]
\centering
    \includegraphics[width=0.6\columnwidth]{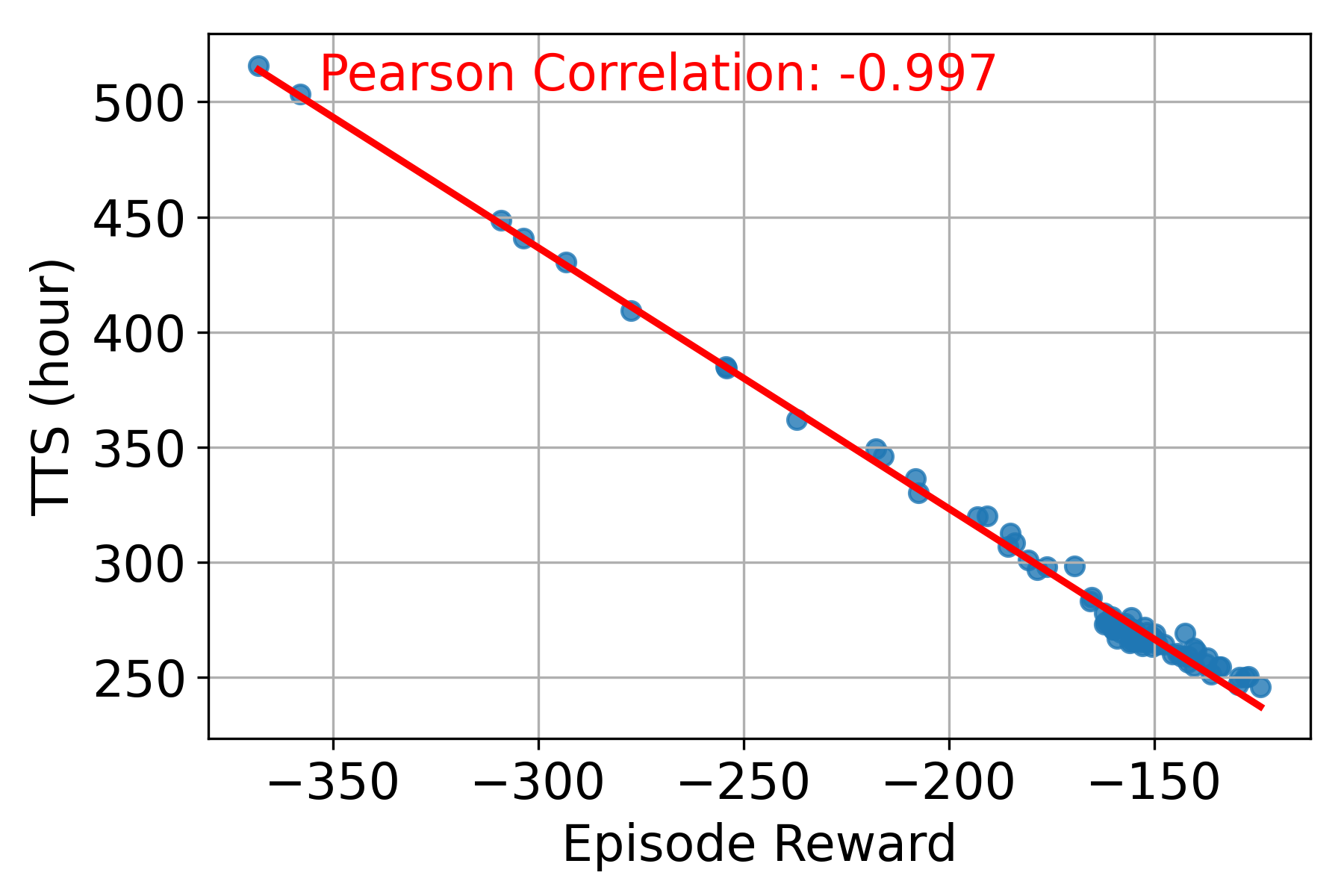}
    \caption{TTS vs Episode Reward.}
    \label{fig:original-vs-surrogate-objective}
\end{figure}

\textit{Learning Diagnosis}: To investigate the relationship between the original objective (TTS) and surrogate objective (episode rewards), Figure \ref{fig:original-vs-surrogate-objective} demonstrates a near-perfect negative linear relationship (Pearson correlation: -0.997) between the episode reward defined in Eq. (\ref{eq:potential-reward}), and TTS in the network. This strong negative correlation indicates that maximizing the episode reward effectively leads to substantial reductions in TTS, thereby validating the use of multi-hop potentials as a reliable surrogate for traffic efficiency. Minimization of the multi-hop upstream potential indicates less congested upstream traffic.

The performance comparison between the proposed method and baselines are shown in Table \ref{tab:performance-comparison}. The proposed farsighted (1-hop) agent beats the pre-timed Webster method and the myopic RL method PressLight, under all demand levels. Webster method has the worst performance as it is not adaptive to the dynamic demand. The performance of PressLight is similar but slightly worse compared to our method with 0-hop upstream in undersaturated and heavily saturated demand levels, because PressLight -- with pressure-based reward -- would deliberately hold vehicles upstream to achieve the minimization of pressure, whereas our potential-based reward encourages vehicles to move downstream.

\begin{figure}[htbp]
\centering
\includegraphics[width=0.6\columnwidth]{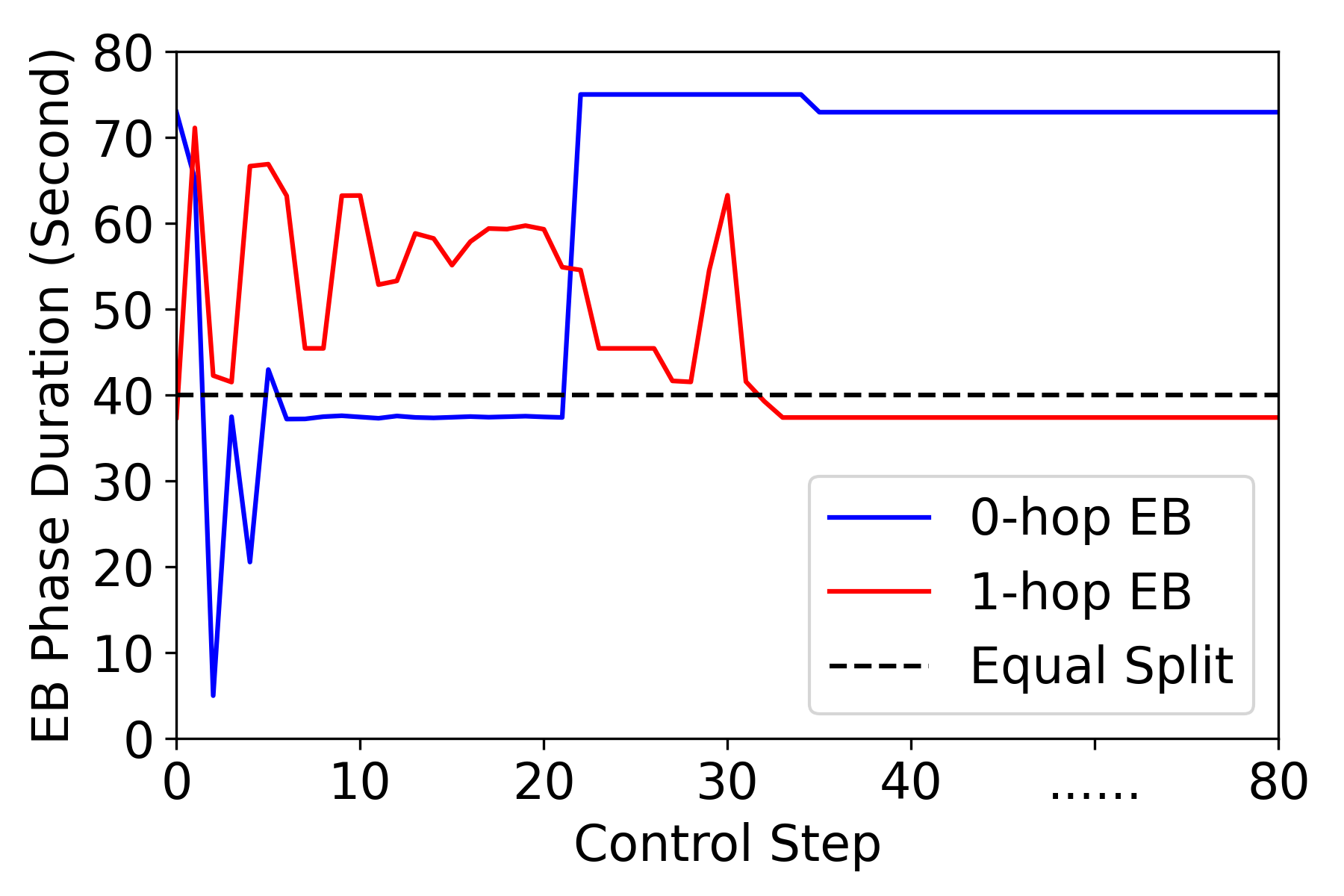}
\caption{The allocated green time for eastbound phase for the right intersection in Network 1x2. The farsighted agent assigns longer green time for eastbound phase when queues exist before time step 30.}
\label{fig:network-1x2-action-vs-time}
\end{figure}

\begin{table*}[!htbp]
\captionsetup{justification=centerlast} 
\centering
\caption{Performance comparison of all methods on synthetic networks}
\label{tab:performance-comparison}
\begin{tabular}{l|ccccc}
\hline
\textbf{Network} & \textbf{Demand Level} & \textbf{Method} & \textbf{TTS} & \textbf{\begin{tabular}[c]{@{}c@{}}Total Queue Time\\ (Include Virtual)\end{tabular}} & \textbf{\begin{tabular}[c]{@{}c@{}}Total Virtual \\Queue Time\end{tabular}} \\ \hline
\multirow{12}{*}{Network 1x2} & \multirow{4}{*}{Undersaturated}     & Webster                 & 21.6           & 6.0              & 0.0                      \\
                              &                                     & PressLight              & 21.0           & 5.5              & 0.0                      \\
                              &                                     & Ours: 0-hop          & 20.8           & 5.3              & 0.0                      \\
                              &                                     & \textbf{Ours: 1-hop}          & \textbf{20.7}  & \textbf{5.2}     & \textbf{0.0}             \\ \cline{2-6} 
                              & \multirow{4}{*}{Slightly Saturated} & Webster                 & 114.8          & 68.4             & 23.5                     \\
                              &                                     & PressLight              & 91.5          & 37.3             & 13.6                     \\
                              &                                     & Ours: 0-hop          & 88.6          & 34.8             & 15.1                     \\
                              &                                     & \textbf{Ours: 1-hop}         & \textbf{76.7} & \textbf{36.7}    & \textbf{1.7}             \\ \cline{2-6} 
                              & \multirow{4}{*}{Heavily Saturated}  & Webster                 & 270.1          & 196.8            & 130.9                    \\
                              &                                     & PressLight              & 249.9          & 197.9            & 145.5                    \\
                              &                                     & Ours: 0-hop          & 242.6          & 187.6            & 138.0                    \\
                              &                                     & \textbf{Ours: 1-hop} & \textbf{221.5} & \textbf{155.8}   & \textbf{64.4}            \\ \hline
\multirow{15}{*}{Network 1x3} & \multirow{5}{*}{Undersaturated}     & Webster                 & 33.6           & 7.86             & 0.0                      \\
                              &                                     & PressLight              & 30.8           & 5.82             & 0.0                      \\
                              &                                     & Ours: 0-hop          & 30.1           & 5.17             & 0.0                      \\
                              &                                     & Ours: 1-hop          & 30.07          & 5.15             & 0.0                      \\
                              &                                     & \textbf{Ours: 2-hop}          & \textbf{30.04} & \textbf{5.12}    & \textbf{0.0}             \\ \cline{2-6} 
                              & \multirow{5}{*}{Slightly Saturated} & Webster                 & 134.4          & 62.6             & 24.4                     \\
                              &                                     & PressLight              & 99.7           & 30.6             & 10.6                     \\
                              &                                     & Ours: 0-hop          & 104.0          & 32.5             & 12.6                     \\
                              &                                     & Ours: 1-hop          & 93.4           & 24.7             & 7.4                      \\
                              &                                     & \textbf{Ours: 2-hop}          & \textbf{85.0}  & \textbf{20.5}    & \textbf{2.7}             \\ \cline{2-6} 
                              & \multirow{5}{*}{Heavily Saturated}  & Webster                 & 325.2          & 210.1            & 132.7                    \\
                              &                                     & PressLight              & 311.1          & 219.1            & 144.5                    \\
                              &                                     & Ours: 0-hop          & 309.3          & 222.9            & 140.4                    \\
                              &                                     & Ours: 1-hop          & 293.7          & 199.9            & 134.4                    \\
                              &                                     & \textbf{\textbf{Ours: 2-hop}} & \textbf{272.9} & \textbf{173.3}   & \textbf{78.9}            \\ \hline
\end{tabular}
\end{table*}

Figure \ref{fig:network-1x2-action-vs-time} illustrates the control action (eastbound phase duration) over time for the right intersection in the Network 1x2 scenario. The southbound green time is automatically determined based on the remaining cycle time, as the total cycle length is fixed. The agent informed by myopic upstream pressures (0 hop) learned near equal splits for the eastbound and southbound phases, as the agent only observed equal queue lengths on immediate eastbound and southbound links, resulting in suboptimal green time allocation. In contrast, agents informed by farsighted pressures (1 hop) as observations and rewards learn better control policies, consistently assigning greater green time to the eastbound phase, with an average of 2.7:1 green time splits for the eastbound and southbound phases, which is crucial for managing queues efficiently in this scenario.

\begin{figure}[htbp]
\centering
\includegraphics[width=\columnwidth]{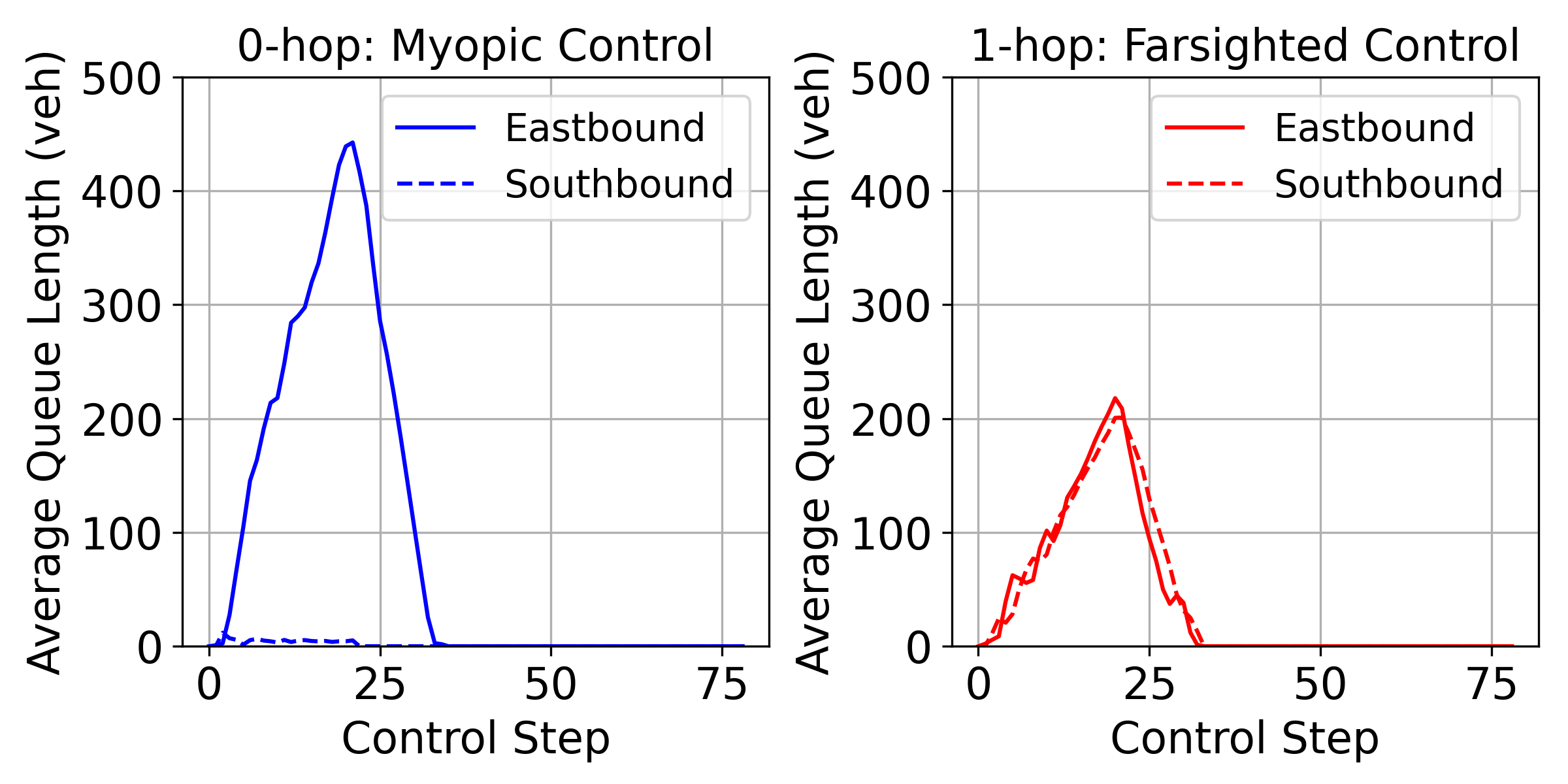}
\caption{Comparison of average queue lengths over time between myopic (0-hop) and farsighted (1-hop) control.}
\label{fig:network-1x2-aqlvq-vs-time}
\end{figure}

Consistent with the control action in Figure \ref{fig:network-1x2-action-vs-time}, the queue lengths under the myopic and farsighted agents are compared in Figure \ref{fig:network-1x2-aqlvq-vs-time}. The myopic agent results in a sharp increase in the queue length on the eastbound. In contrast, the farsighted agent has lower peak queue lengths and more balanced eastbound and southbound queue lengths. The sudden increase of eastbound phase duration for the myopic agent is because no new vehicles are generated after timestep 20, and southbound queues are cleared at timestep 22, thus the myopic agent can finally assign maximal allowed green time to clear eastbound queues. This comparison highlights the advantage of multi-hop upstream pressures and potentials into the RL agent design that significantly reduce congestion in tested scenarios.

\subsection{Synthetic Scenario: Network 1x3}

\begin{figure*}[htbp]
\centering
  \begin{subfigure}[b]{0.66\columnwidth}
    \includegraphics[width=\linewidth]{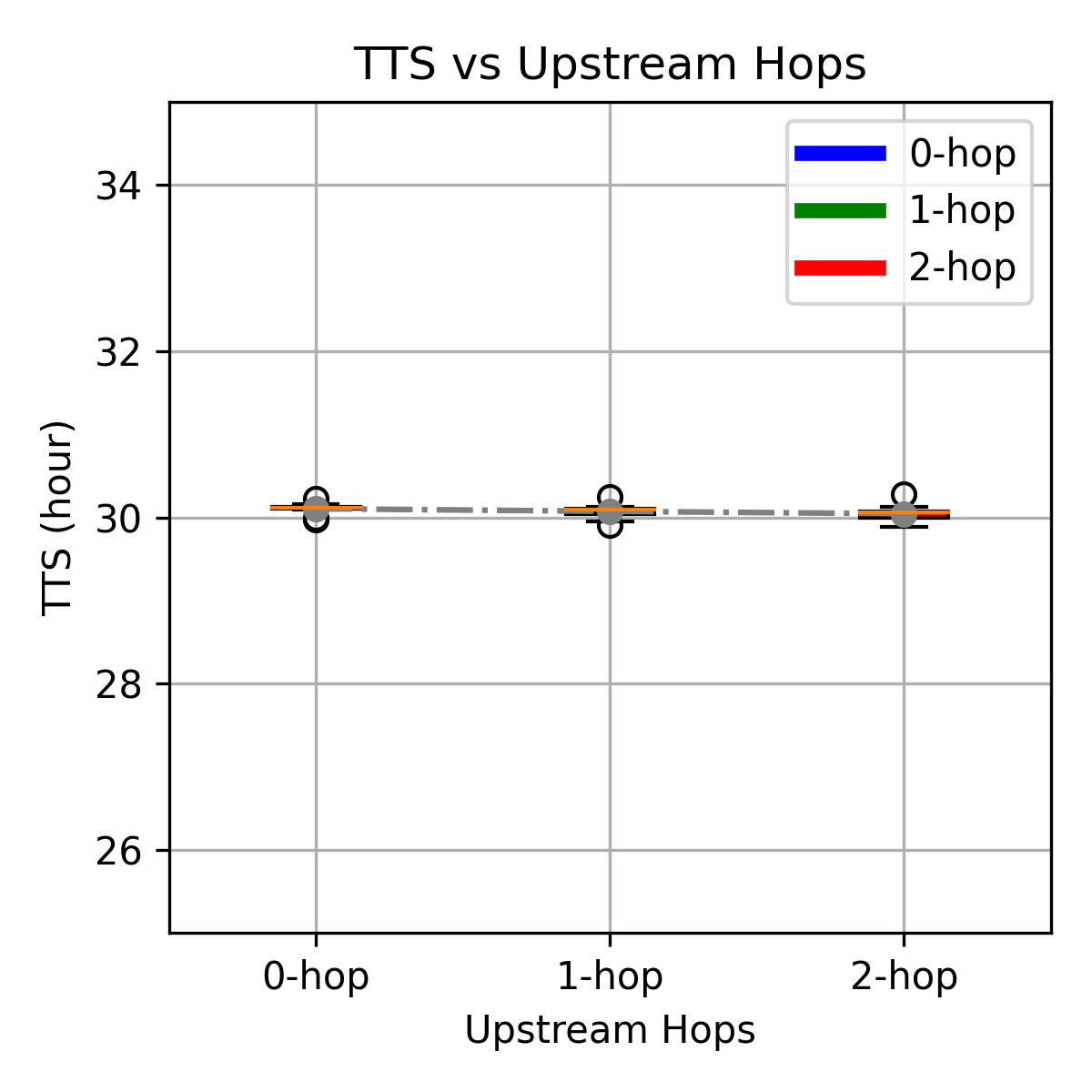}
    \caption{Undersaturated demand.}
    \label{fig:tts-vs-undersaturated-demand}
  \end{subfigure}
  \begin{subfigure}[b]{0.66\columnwidth}
    \includegraphics[width=\linewidth]{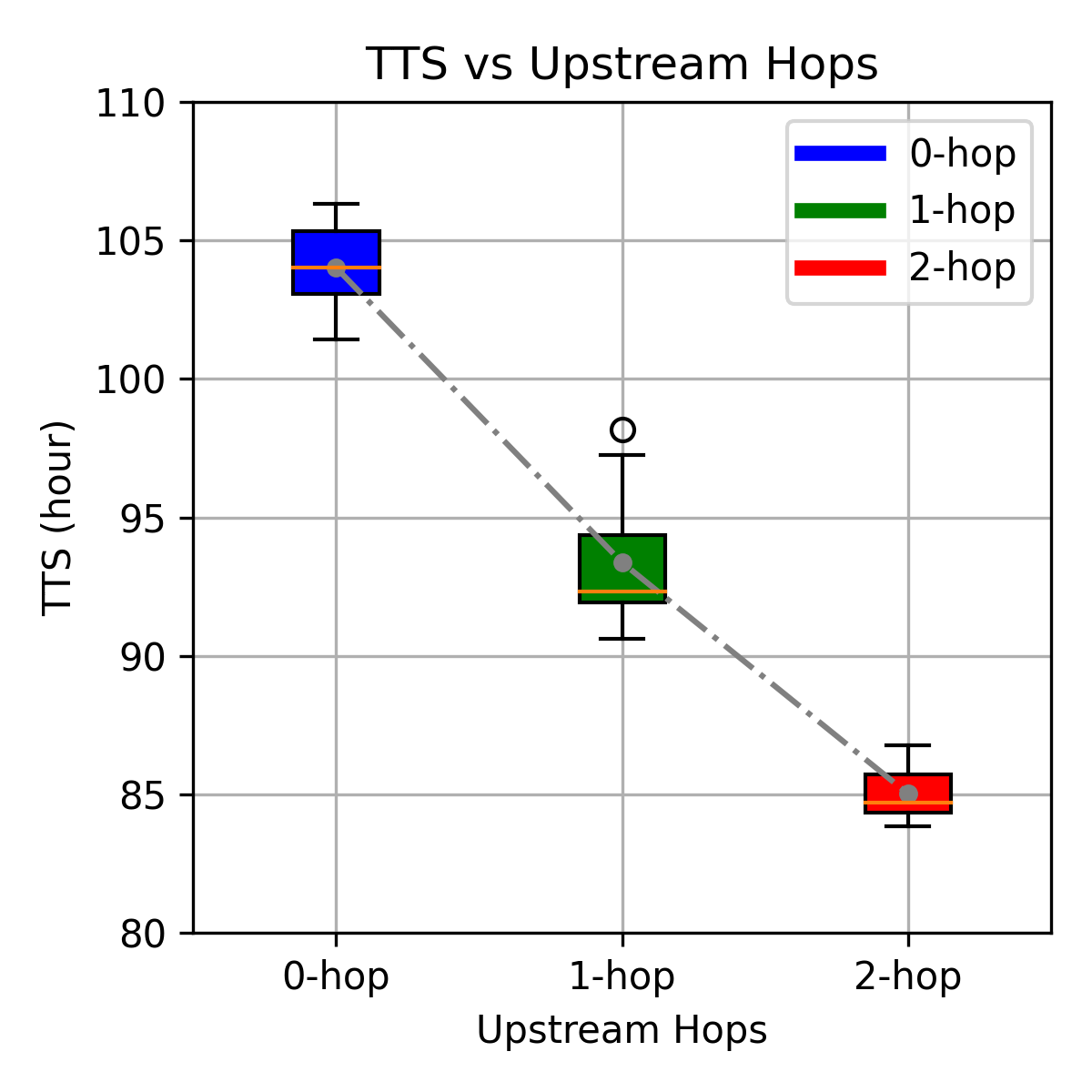}
    \caption{Slightly saturated demand.}
    \label{fig:tts-vs-slightly-saturated-demand}
  \end{subfigure}
  \begin{subfigure}[b]{0.66\columnwidth}
    \includegraphics[width=\linewidth]{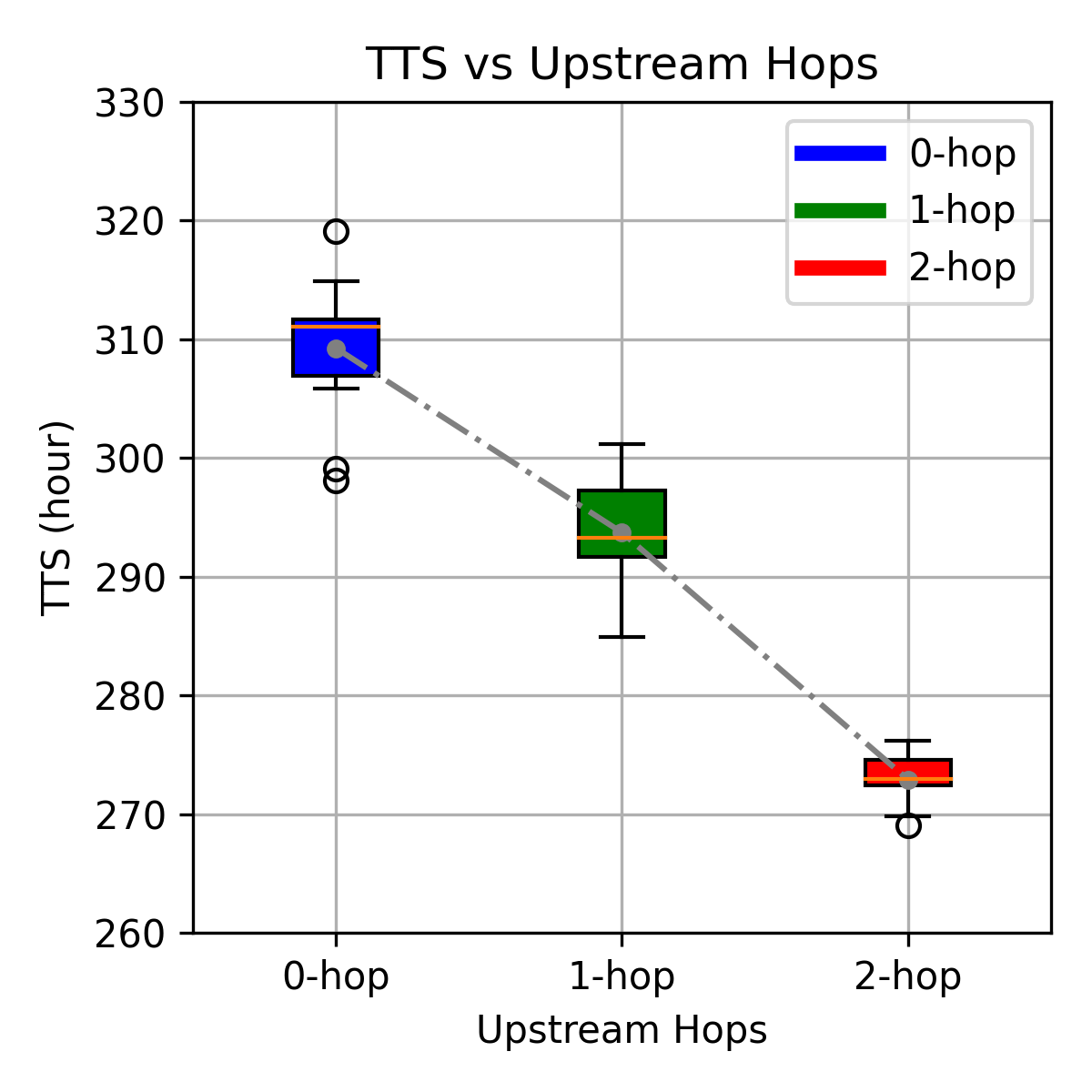}
    \caption{Heavily saturated demand.}
    \label{fig:tts-vs-heavily-saturated-demand}
  \end{subfigure}
\caption{TTS vs upstream hops for (a) \textit{undersaturated} demand, (b) \textit{slightly} saturated demand, (c) \textit{heavily} saturated demand. Farsighted pressures contribute to an improved performance when the demand is saturated, and does no harm to undersaturated cases.}
\label{fig:tts-vs-hop}
\end{figure*}

Figure \ref{fig:tts-vs-hop} presents the impact of upstream hop information on total time spent (TTS) under three demand levels: (a) undersaturated, (b) slightly saturated, and (c) heavily saturated:
\begin{itemize}
    \item In undersaturated conditions (Figure \ref{fig:tts-vs-undersaturated-demand}), TTS remains nearly constant across 0-hop, 1-hop, and 2-hop scenarios, with minimal variation. This indicates that in low-demand conditions, the inclusion of additional upstream hop information has no adverse impact on performance. Since the network is not congested, simpler control using immediate pressures (0-hop) is sufficient to achieve optimal performance.
    \item In slightly saturated conditions (Figure \ref{fig:tts-vs-slightly-saturated-demand}), TTS begins to show variation across the different hop levels. The 1-hop and 2-hop configurations achieve lower TTS than the 0-hop scenario, with the 2-hop configuration yielding the lowest TTS. This suggests that under moderate congestion, farsighted setups (1-hop and 2-hop) enable the agents to coordinate more effectively, reducing delays by considering upstream traffic conditions.
    \item In heavily saturated conditions (Figure \ref{fig:tts-vs-heavily-saturated-demand}), the benefits of using farsighted pressure still exist. The TTS decreases progressively from 0-hop to 2-hop. This demonstrates that when the network is heavily congested, incorporating additional upstream information better manages the traffic.
\end{itemize}

The results indicate that farsighted pressure (1-hop and 2-hop) provides substantial benefits in saturated and heavily saturated scenarios by enabling more effective green time allocation based on upstream congestion levels. In contrast, in undersaturated scenarios, the additional upstream information does not affect performance, as immediate pressures alone suffice to maintain optimal flow. Therefore, multi-hop pressure control improves traffic efficiency for saturated conditions without detriment to undersaturated cases.

\subsection{Realistic Scenario: Toronto Testbed}

In this section, we discuss the results for the Toronto testbed.

\begin{figure}[htbp]
\centering
\includegraphics[width=\columnwidth]{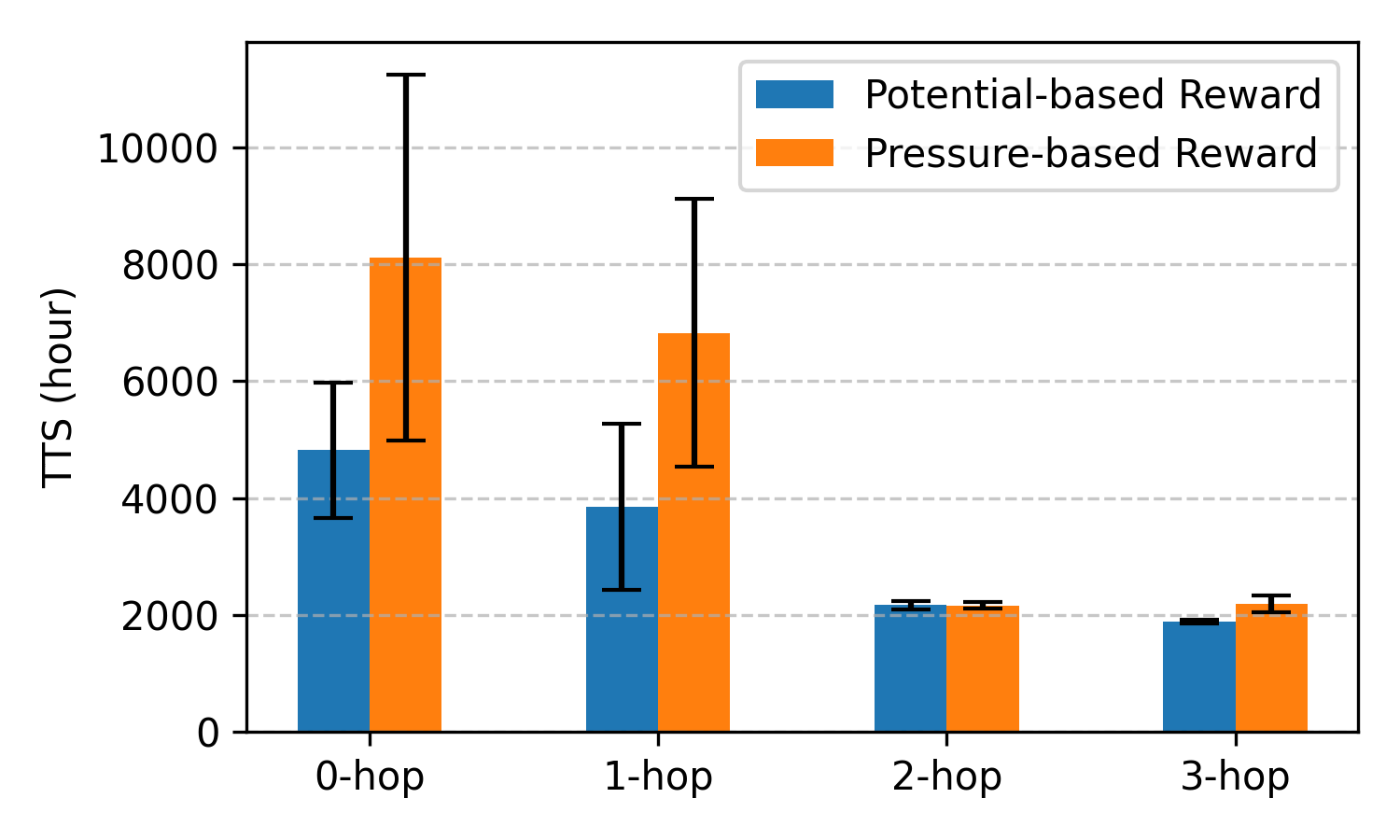}
\caption{
Potential-based vs pressure-based reward design.}
\label{fig:potential-vs-pressure-reward}
\end{figure}

Figure \ref{fig:potential-vs-pressure-reward} compares the performance of potential-based reward and the pressure-based reward, previously defined in Eq. (\ref{eq:potential-reward}) and Eq. (\ref{eq:pressure-reward}). Although both PressLight (0-hop pressure-based reward) and our 0-hop potential-based RL method are myopic, PressLight exhibits higher variability (less robustness) in performance. This may be attributed to its pressure-based reward design, which can unintentionally encourage the agent to hold vehicles upstream to minimize local pressure, leading to higher TTS and instability. The higher variability of RL agents using pressure-based rewards has also been observed in empirical evaluations \cite{jayawardana2022impact, zhang2023evaluation}. In contrast, our potential-based reward consistently incentivizes the agent to release vehicles to downstream links, promoting smoother traffic flow and reducing variability. \textit{In later results and discussions, we only focus on potential-based reward.}

\begin{figure}[htbp]
\centering
\includegraphics[width=\columnwidth]{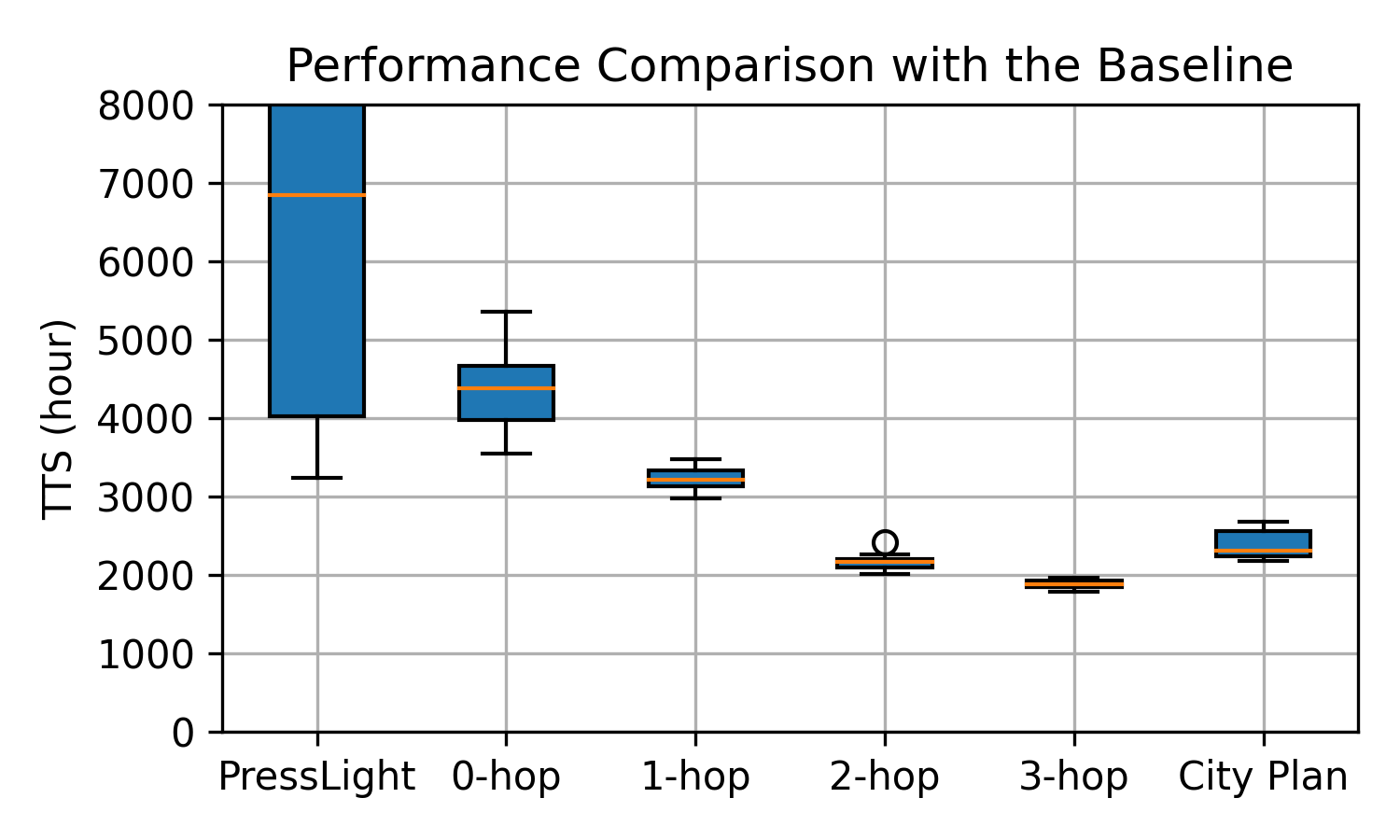}
\caption{
Performance comparison between our methods and the baseline. The variability comes from 10 different replications by setting 10 unique random seeds.}
\label{fig:tts-toronto}
\end{figure}

Figure \ref{fig:tts-toronto} compares the TTS among our methods and the City Plan baseline on the Toronto testbed, cross validated by queue time comparison in Table \ref{tab:performance-toronto-testbed}. Compared to the City Plan (2303 hours), Our RL models using 0-hop and 1-hop upstream setups show higher TTS (4383 and 3214 hours, respectively). The increased total and virtual queue times indicate that these models struggle due to their myopic observation and reward. In contrast, the 2-hop and 3-hop RL models outperform the City Plan, with 6\% and 19\% improvement on TTS, respectively. The results demonstrated the benefit of incorporating multi-hop upstream traffic, enabling a more coordinated signal control.

\begin{table}[!htbp]
\captionsetup{justification=centerlast}
\centering
\caption{Average performance on the Toronto testbed}
\label{tab:performance-toronto-testbed}
\begin{tabular}{cccc}
\hline
\textbf{Method} & \textbf{TTS} & \textbf{\begin{tabular}[c]{@{}c@{}}Total Queue Time\\ (Include Virtual)\end{tabular}} & \textbf{\begin{tabular}[c]{@{}c@{}}Total Virtual\\ Queue Time\end{tabular}}  \\ \hline
City Plan  & 2303          & 1312         & 367        \\ \hline
PressLight          & 8116          & 7501         & 2809       \\ \hline
Ours: 0-hop          & 4383          & 2640         & 1110       \\ \hline
Ours: 1-hop          & 3214          & 2032         & 83         \\ \hline
Ours: 2-hop          & 2170          & 1104         & 89         \\ \hline
\textbf{Ours: 3-hop} & \textbf{1878} & \textbf{880} & \textbf{3} \\ \hline
\end{tabular}
\end{table}

\begin{figure*}[htbp]
\centering
\includegraphics[width=0.95\textwidth]{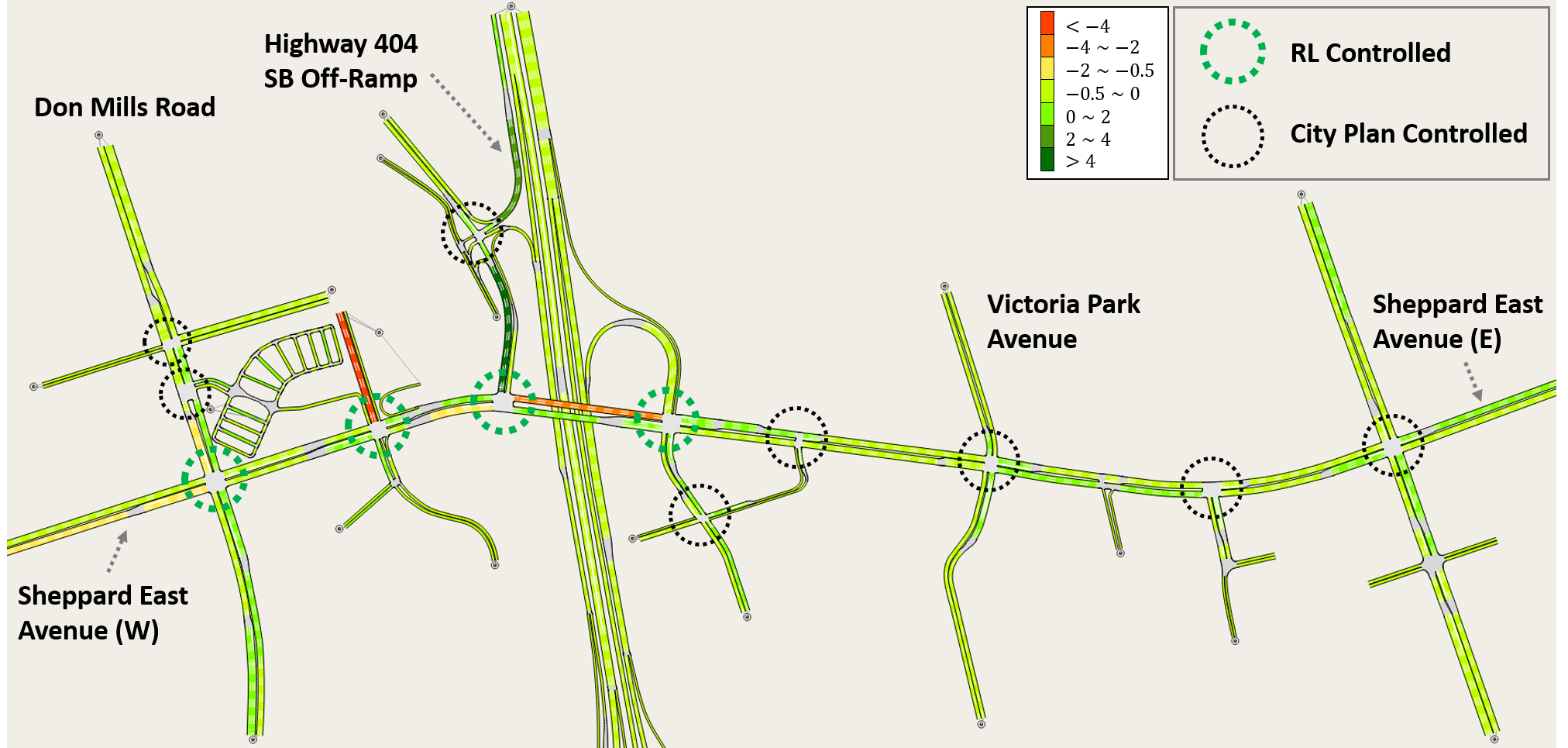}
\caption{The heatmap of queue difference between the City Plan and our RL approach with 3-hop upstream setup. The road-wise comparison is based on the average queue length over the whole simulation time. Roads labeled in green indicate improved queue length in our approach compared to the City Plan, while roads in red indicate longer queues.}
\label{fig:heatmap-sheppard-hwy404}
\end{figure*}

The heatmap in Figure \ref{fig:heatmap-sheppard-hwy404} highlights the differences in queue lengths between the City Plan and our RL approach using 3-hop upstream setup. The compared queue lengths are average values throughout the whole simulation, better reflecting the congestion than just comparing on a time slice. Green roads indicate reduced queue lengths with the RL approach, while red roads show increased queues. 

\begin{table}[!htbp]
\captionsetup{justification=centerlast}
\centering
\caption{Top 4 largest number of trips origins.}
\label{tab:top-4-origin}
\begin{tabular}{cccc}
\hline
\textbf{Origin}      & \textbf{Trips (\%)}  & \textbf{Trips (veh)}    \\ \hline
Highway 404 N        & 15.5\%   &  4241         \\ \hline
Don Mills Road N          & 12.5\%   &  3415         \\ \hline
Don Mills Road S          & 9.4\%    &  2579        \\ \hline
Sheppard Avenue W       & 9.3\%        &  2555  \\ \hline
\end{tabular}
\end{table}

A key improvement is observed at the Highway 404 southbound off-ramp. The RL approach focuses on optimizing the Highway 404 off-ramp, which is supported by the trip distribution data in Table \ref{tab:top-4-origin}
that 15.5\% of local trips originate from the Highway 404 North origin, which is also the largest trip numbers, making it a critical point for signal control. Overall, most roads benefit from our proposed approach, indicating improved traffic flow. While the southbound link connected to the Fairview Mall (second left intersection) and the westbound link (third left intersection) experience increased queues as a result of competition among different signal phases, these increased queues are compensated by the significant improvement in prioritizing the Highway 404 SB off-ramp. This farsighted prioritization obtains a substantial gain rather than merely a trade-off.

Our RL approach also learns intersection coordination: traffic from the Highway 404 off-ramp, which feeds into Sheppard Avenue, requires adjacent intersections (the second and fourth RL-controlled intersections counting from the left) to coordinate by allocating longer green times along the arterial road to manage the incoming flow effectively. Our RL approach achieved this coordination, as the two adjacent RL-controlled intersection's incoming links are greatly improved on queues, rendered in  green. 


\section{CONCLUSION}
This paper introduces a novel concept of multi-hop upstream pressure and integrates it to RL agent design to address the limitations of myopic pressure-based control methods that focus solely on immediate upstream links. The proposed multi-hop upstream pressure accounts for an abstracted view over a greater upstream area beyond the immediate upstream link, providing a broader spatial awareness for optimizing signal timings and achieving coordination. 



Our experiments, conducted in both the synthetic scenario and the realistic Toronto testbed scenario, demonstrate that the RL agents utilizing the multi-hop upstream metric perform better in reducing network delays compared to myopic approaches. Notably, the approach performs exceptionally well in oversaturated scenarios and remains effective in undersaturated scenarios, benefiting from multi-hop information from further upstream links.

Future work could refine this approach by incorporating dynamic turning ratio estimation, expanding the scope to more complex networks, and applying multi-hop pressure to other traffic control problems, such as congestion pricing and dynamic perimeter identification.

\section*{APPENDIX}
\section*{Scalar Version: Multi-hop Upstream Pressure for a Single Link}

To gently guide the reader step-by-step, we first review the vanilla version of traffic pressure, then demonstrate how to extend it to higher-hop upstream versions.

\paragraph{Pressure with 0-hop Upstream:} Adopted from physics, the existing standard traffic pressure \cite{varaiya2013maxpressure} is defined as the difference between immediate upstream queue length and the summation of immediate downstream queue lengths weighted by turning ratios. Mathematically, for a link $l$, its pressure with 0-hop upstream is:
\begin{align}
p(l, 0) = Q(l) - \sum_{j \in \mathcal{N}_d (l,1)} T_{lj} Q(j)
\end{align}
Apparently, 0-hop upstream is myopic in terms of knowing the traffic conditions beyond the link of interest. When considering further neighborhoods, the concept of 0-hop upstream can be extended to multi-hop upstream to account for a wider range of traffic networks, capturing the cumulative effect of traffic congestion in neighboring areas.

\paragraph{Pressure with 1-hop Upstream:} Compared to pressure with 0-hop upstream, extra traffic information from 1-hop upstream links is integrated. The influence of 1-hop upstream links on the current link $l$ is naturally weighted by the turning ratio from 1-hop upstream links to link $l$:
\begin{align}
    p(l,1) &= p(l,0) + \sum_{i_1 \in \mathcal{N}_u (l,1)} T_{i_1 l} Q(i_1) \\
    &=  p(l,0) + (\mathbf{P}_{:, l})^\top\mathbf{Q}
\end{align}

\paragraph{Pressure with 2-hop Upstream:} Similarly, the impact of 2-hop upstream links is added upon pressure with 1-hop upstream. The congestion at 2-hop upstream links has less influence than 1-hop upstream links on the current link $l$, and is naturally discounted by the turning ratio from 2-hop upstream links to link $l$:
\begin{align}
    p(l,2) &= p(l,1) + \sum_{i_1 \in \mathcal{N}_u (l,1)} \sum_{i_2 \in \mathcal{N}_u (i_1,1)}  T_{i_2 i_1} T_{i_1 l} Q(i_2) \\
    &=  p(l,1) +  [(\mathbf{P}^2)_{:, l}]^\top\mathbf{Q}
\end{align}

\paragraph{Pressure with $h$-hop Upstream:} To generalize, the impact of $h$-hop upstream links is added upon pressure with $(h-1)$-hop upstream. The decay of influence from $h$-hop upstream links to link $l$ is captured by the turning ratio from $h$-hop upstream links to link $l$:
\begin{align}
    p(h,1) &= p(l,h-1) + \sum_{i_1 \in \mathcal{N}_u(l, 1)} \sum_{i_2 \in \mathcal{N}_u (i_1, 1)} ... \sum_{i_h \in \mathcal{N}_u (i_{h-1}, 1)} \nonumber \\ & \phantom{p(h,1) = p(l,h-1)} T_{i_h i_{h-1}}...T_{i_2 i_1} T_{i_1 l}Q(i_h) \\
    &=  p(l,h-1) +  [(\mathbf{P}^h)_{:, l}]^\top\mathbf{Q}
\end{align}

\section*{ACKNOWLEDGMENT}

\bibliographystyle{IEEEtran}
\bibliography{refs}

\begin{IEEEbiography}[{\includegraphics[width=1in,height=1.25in,clip,keepaspectratio]{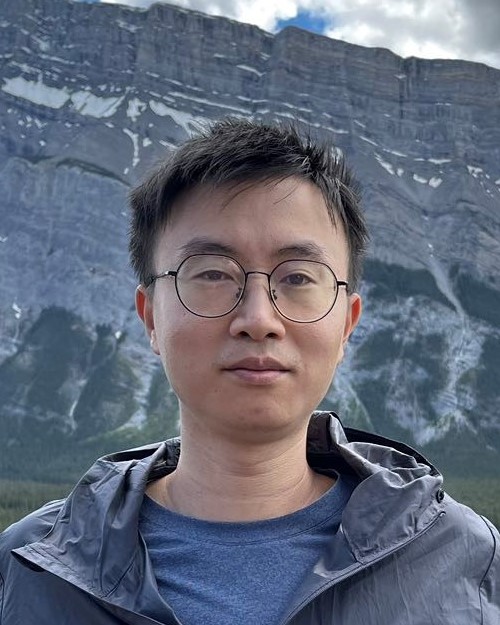}}]{Xiaocan Li } received the B.Eng. degree from Beihang University, Beijing, China, in 2017, and the M.Sc. degree in Control Theory and Engineering from the Institute of Automation, Chinese Academy of Sciences, Beijing, China, in 2020. He is currently a Ph.D. candidate with the Department of Mechanical \& Industrial Engineering at the University of Toronto, ON, Canada. His research interests include deep reinforcement learning, spatiotemporal prediction, and their application to intelligent transportation systems.
\end{IEEEbiography}

\begin{IEEEbiography}[{\includegraphics[width=1in,height=1.25in,clip,keepaspectratio]{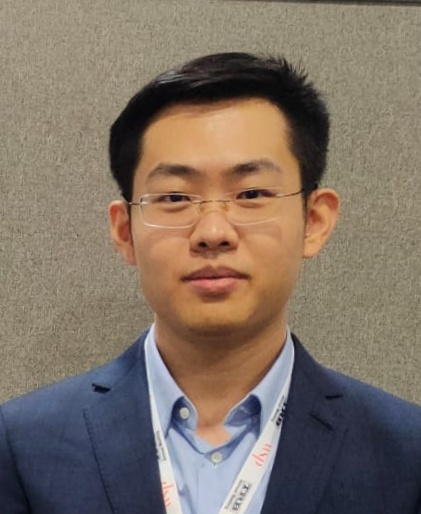}}]{Xiaoyu Wang } received the B.Eng. degree in automation from Tianjin University, Tianjin, China, in 2016, and the M.Sc. degree in control science and technology from Shanghai Jiao Tong University, Shanghai, China, in 2019. He is currently a Ph.D. candidate with the Department of Civil and Mineral Engineering at the University of Toronto, ON, Canada. His research interests include control, reinforcement learning, and their application to intelligent transportation and multi-agent systems.
\end{IEEEbiography}

\begin{IEEEbiography}[{\includegraphics[width=1in,height=1.25in,clip,keepaspectratio]{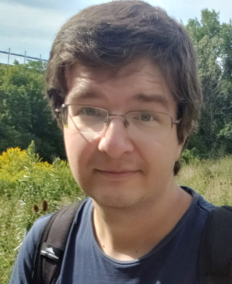}}]{Ilia Smirnov } is a Research Associate at the Department of Civil and Mineral Engineering at the University of Toronto. He completed a Ph.D. in Pure Mathematics (Algebraic Geometry) at Queen's University, ON, Canada in 2020. His research interests include Planning, Reinforcement Learning, and Optimization in Intelligent Transportation Systems, as well as Enumerative Algebraic Geometry / Intersection Theory.
\end{IEEEbiography}

\begin{IEEEbiography}[{\includegraphics[width=1in,height=1.25in,clip,keepaspectratio]{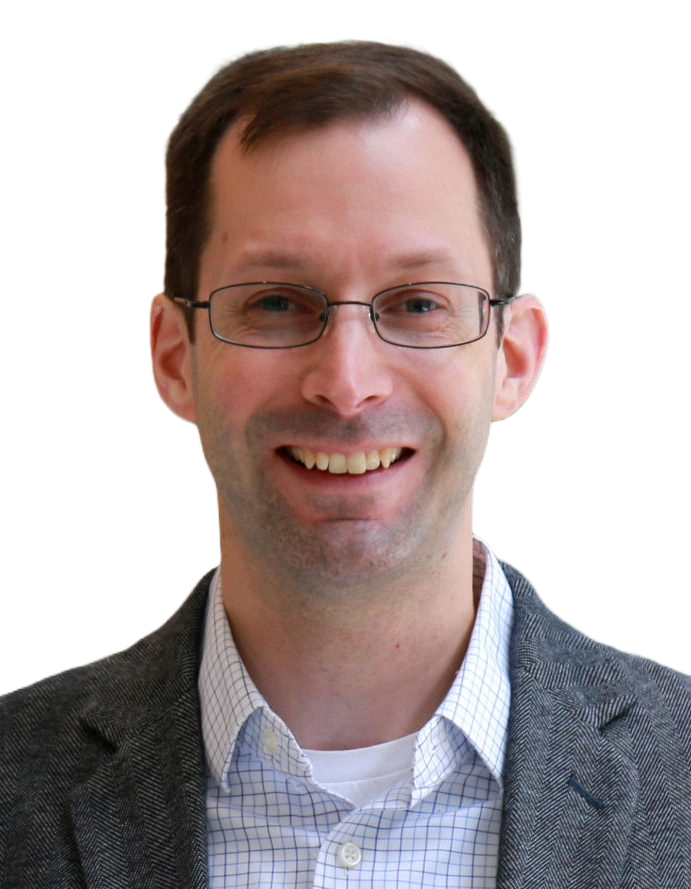}}]{Scott Sanner } received the B.Sc. degree in computer science from the Carnegie Mellon University, Pittsburgh, PA, USA, in 1999, the M.Sc. degree in computer science from Stanford University, Stanford, CA, USA, in 2002, and the Ph.D. degree in computer science from the University of Toronto, ON, Canada, in 2008. He is an Associate Professor in Industrial Engineering and Cross-appointed in Computer Science at the University of Toronto. Scott’s research focuses on a broad range of AI topics spanning sequential decision-making and applications of machine/deep learning to Smart Cities. Scott is currently an Associate Editor for the Machine Learning Journal (MLJ) and the Journal of Artificial Intelligence Research (JAIR). Scott was a co-recipient of paper awards from the AI Journal (2014), Transport Research Board (2016), CPAIOR (2018) and a recipient of a Google Faculty Research Award in 2020.
\end{IEEEbiography}

\begin{IEEEbiography}[{\includegraphics[width=1in,height=1.25in,clip,keepaspectratio]{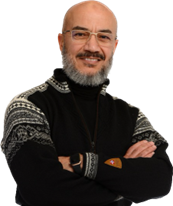}}]{Baher Abdulhai } received the Ph.D. degree in engineering from the University of California, Irvine, CA, USA, in 1996. He is a Professor in Civi Engineering at the University of Toronto, ON, Canada. He has 35 years of experience in transportation systems engineering and Intelligent Transportation Systems (ITS). He is the founder and Director of the Toronto Intelligent Transportation System Center, and the founder and co-Director of the i-City Center for Automated and Transformative Transportation Systems (iCity-CATTS).
He received several awards including IEEE Outstanding Service Award, Teaching Excellence award, and research awards from Canada Foundation for Innovation, Ontario Research Fund, and Ontario Innovation Trust.  He served on the Board of Directors of the Government of Ontario (GO) Transit Authority from 2004 to 2006.  He served as a Canada Research Chair (CRC) in ITS from 2005 to 2010. His research team won international awards including the International Transportation Forum innovation award in 2010 (Hossam Abdelgawad), IEEE ITS 2013 (Samah El-Tantawy) and INFORMS 2013 (Samah El-Tantawy). In 2015 he has been inducted as a Fellow of the Engineering Institute of Canada (EIC). In 2018, he won the prestigious CSCE Sandford Fleming (Career Achievement) Award for his contribution to transportation in Canada. He has been elected Fellow of the Canadian Academy of Engineering in 2020. In 2021, he won the Ontario Professional Engineers Awards (OPEA) Engineering Medal for career Engineering Excellence.
\end{IEEEbiography}

\end{document}